%% file: main.tex
\documentclass{article}
\usepackage[numbers,sort,compress]{natbib}
\usepackage[breaklinks,colorlinks]{hyperref}

\usepackage[final]{neurips_2022}
\usepackage[utf8]{inputenc} % allow utf-8 input
\usepackage[T1]{fontenc}    % use 8-bit T1 fonts
\usepackage{hyperref}       % hyperlinks
\usepackage{url}            % simple URL typesetting
\usepackage{booktabs}       % professional-quality tables
\usepackage{amsfonts}       % blackboard math symbols
\usepackage{nicefrac}       % compact symbols for 1/2, etc.
\usepackage{microtype}      % microtypography
\usepackage{xcolor}         % colors
\usepackage{url}
\usepackage{booktabs}
\usepackage{graphicx}
\usepackage{xspace}
\usepackage{subcaption}
\usepackage{makecell}
\usepackage{pifont}
\usepackage{amssymb}
\usepackage{enumitem}
\usepackage{wrapfig}
\newcommand{\cmark}{\ding{51}}
\newcommand{\xmark}{\ding{55}}

\definecolor{lightgrey}{rgb}{0.43,0.43,0.43}
\definecolor{crimson}{rgb}{0.86,0.08,0.24}
\hypersetup{
  colorlinks,
  citecolor=lightgrey,
  linkcolor=crimson,
  urlcolor=lightgrey}

\definecolor{darkgreen}{rgb}{0,0.39,0}

\definecolor{deepmagenta}{rgb}{0.8, 0.0, 0.8}

\definecolor{darkorange}{rgb}{0.85, 0.37, 0.01}

\definecolor{blue}{rgb}{0.0, 0.0, 1.0}

\newcommand{\modelname}{SAVi\texttt{+}\texttt{+}\xspace}

\newcommand{\ebar}[2]{$#1 \scriptstyle{\,\pm\,#2}$}

\title{\modelname: Towards End-to-End Object-Centric Learning from Real-World Videos
}

\newcommand\blfootnote[1]{%
  \begingroup
  \renewcommand\thefootnote{}\footnote{#1}%
  \addtocounter{footnote}{-1}%
  \endgroup
}

\author{Gamaleldin F.~Elsayed$^{\,*\dagger}$, Aravindh Mahendran$^{*\diamond}$, Sjoerd van Steenkiste$^{*\diamond}$, \\\textbf{Klaus Greff, Michael C.~Mozer \& Thomas Kipf$^*$}\\
Google Research}

\begin{document}

\maketitle

\begin{abstract}
The visual world can be parsimoniously characterized in terms of distinct entities with sparse interactions. Discovering this compositional structure in dynamic visual scenes has proven challenging for end-to-end computer vision approaches unless explicit instance-level supervision is provided. Slot-based models leveraging motion cues have recently shown great promise in learning to represent, segment, and track objects without direct supervision, but they still fail to scale to complex real-world multi-object videos. In an effort to bridge this gap, we take inspiration from human development and hypothesize that information about scene geometry in the form of depth signals can facilitate object-centric learning. We introduce \modelname, an object-centric video model which is trained to predict depth signals from a slot-based video representation. By further leveraging best practices for model scaling, we are able to train \modelname to segment complex dynamic scenes recorded with moving cameras, containing both static and moving objects of diverse appearance on naturalistic backgrounds, without the need for segmentation supervision. Finally, we demonstrate that by using sparse depth signals obtained from LiDAR, \modelname is able to learn emergent object segmentation and tracking from videos in the real-world Waymo Open dataset.

Project page: \href{https://slot-attention-video.github.io/savi++/}{\small\texttt{https://slot-attention-video.github.io/savi++/}}
\end{abstract}
\blfootnote{\kern-1.7em$^*$Equal technical contribution. $^\diamond$Alphabetical order. $^\dagger$Correspondence to: \href{mailto:gamaleldin@google.com}{\texttt{gamaleldin@google.com}}\\
Author contributions: GFE, TK initiated and led the project. GFE, AM, SVS, TK developed the main model. GFE, AM, TK developed real-world driving data and model infrastructure. AM implemented data augmentation. GFE led ablation study. SVS led target signals analyses and metric design. GFE, AM, SVS, TK ran experiments. AM, SVS, KG, TK worked on baselines. KG, MCM provided advice at all stages and helped with project scoping. TK developed visualizations. GFE, MCM, TK worked on figure design. All authors wrote the paper.
}
\input{introduction}

\input{related_work}
\input{methods}

\input{experiments}
\input{conclusion}

\section{Acknowledgements}
We would like to thank Ben Caine, Alex Bewley and Pei Sun for assistance with self-driving data. We are grateful to Jie Tan, Daniel Keysers, David Fleet, Matthias Minderer, Mehdi Sajjadi and Mario Lučić for general advice and feedback.

{
    \small
    \bibliographystyle{plainnat}
    \bibliography{savi_pp}
}

\newpage
\input{appendix}

\end{document}

%% file: introduction.tex
\section{Introduction}
\label{sec: introduction}

\begin{wrapfigure}{r}{0.58\textwidth}
  \vspace{-50pt}
    \centering
    \includegraphics[width=\linewidth]{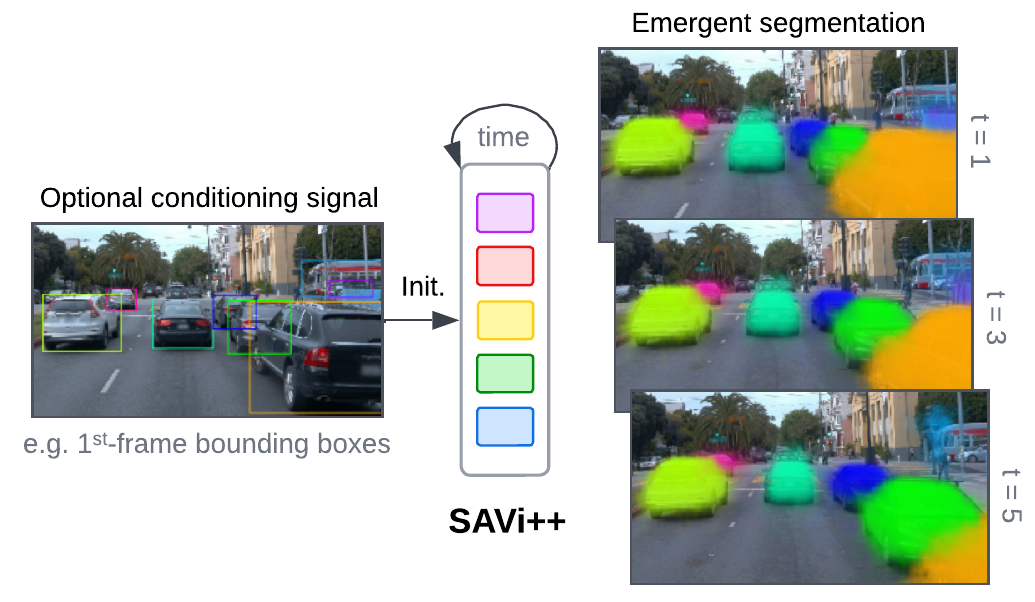}
    \caption{Emergent segmentation and tracking in \modelname.}
    \label{fig:teaser_figure}
    \vspace{-1.5em}
\end{wrapfigure}

The natural world consists of distinct entities---people, dogs, cars, trees, etc.---and its complexity emerges 
from the combined, mostly independent, actions of the entities. This compositional structure must be appreciated
to predict future states of the world and to effect particular outcomes. People have an intrinsic 
understanding of objects: objects have spatiotemporal coherence, they interact when in close 
proximity, and they possess persistent, latent characteristics that determine their behavior over
extended periods of time \citep{kahneman1992reviewing,spelke2007core}. Just as object-centric 
representations are critical to human understanding, they have the potential in machine learning 
to greatly improve sample efficiency, robustness, visual reasoning, and interpretability of 
learning algorithms \citep{greff2020binding,lake2017building}.
For example, consider the challenge faced by
an autonomous vehicle operating in diverse surroundings (Figure~\ref{fig:teaser_figure}). Generalization across
situations requires learning about recurring entities like cars, traffic lights, and pedestrians, and
the rules that govern interactions among these entities. 

In human brains, the ability to organize edges and surfaces into unitary, bounded, and persisting object representations develops through experience and/or maturation from infancy and without explicit instruction via a `core system of object representation' \citep{spelke2007core}, i.e., a form of cognitive inductive bias. In deep learning, such an inductive bias has been proposed in \emph{slot-based} architectures
which segregate knowledge about individual objects into nonoverlapping but interchangeable pools of 
neurons. The resulting representational modularity can facilitate causal reasoning and prediction for 
downstream tasks~\citep{scholkopf2021toward,greff2020binding}.\looseness=-1

A grand challenge in computer vision has been to discover the compositional structure of real-world 
dynamic visual scenes in an unsupervised fashion. 
By unsupervised, we mean no segmentation information is provided that specifies which pixels belong together as part of a single object.
Initial efforts focused on 
single-frame,  synthetic RGB images \citep{greff2017neural,greff2019multi,locatello2020object,van2018relational},
but extending this work to video and more complex scenes proved challenging. A key insight to further
progress was the realization that a color-intensity pixel array is not the only source of visual 
information readily available, at least not to human perceptual systems. The human perceptual system extracts motion and depth cues early in the processing stream
\citep{nakayama1986,enns1990influence,enns1990sensitivity,driver1992,horowitz2007}. These cues
are correlated with object identities, and can therefore bootstrap the formation
of object-centric representations \citep{spelke1990}.

The recently introduced \textit{Slot Attention for Video} (SAVi) model~\citep{kipf2021conditional} leveraged \emph{optical flow} (frame-to-frame motion) as a prediction target to obtain object-centric
representations of dynamic scenes involving complex 3D scanned objects and real-world backgrounds. 
However, motion prediction alone is insufficient to learn about the distinction between static objects and the background.
Further, in real-world application domains such as self-driving cars, cameras themselves are subject to movement, which globally affects frame-to-frame motion as a prediction signal in non-trivial ways.

In the present work, we describe an enhanced slot-based model for video, referred to as 
\emph{SAVi++} (Figure~\ref{fig:savi++}), which obtains qualitative improvements in object-centric representations by 
exploiting \emph{depth} signals readily available from RGB-D cameras and LiDAR sensors.
SAVi++ is the first slot-based, end-to-end trained model that successfully segments complex objects in naturalistic, real-world video sequences without using direct segmentation or tracking supervision.

A summary of our contributions is as follows:
\begin{itemize}[leftmargin=*]
    \item We introduce \modelname: an object-centric slot-based video model that makes several key improvements to SAVi~\citep{kipf2021conditional} by utilizing \emph{depth prediction} and by adopting best practices for model scaling in terms of \emph{architecture design} and \emph{data augmentation}.
    \item On the multi-object video (MOVi) benchmark containing synthetic videos of high visual and dynamic complexity \citep{kubric2021}, we find that \modelname is able to handle videos containing complex shapes and backgrounds, and a large number of objects per scene. Improving on SAVi, our approach accommodates both static and dynamic objects and both static and moving cameras.
    \item Finally, we demonstrate that \modelname trained with sparse depth signals obtained from LiDAR enables emergent object decomposition and tracking in real-world driving videos from the Waymo Open dataset~\citep{sun2020scalability}.
\end{itemize}

%% file: related_work.tex
\section{Related work}
\label{sec: related work}

\textbf{Object-centric learning}\quad
A growing body of research is addressing the problem of end-to-end learning of object-centric representations from raw perceptual data without direct supervision. Slot-based neural networks such as IODINE~\citep{greff2019multi}, MONet~\citep{burgess2019monet}, and Slot Attention~\citep{locatello2020object} rely on a factorized latent space and independent per-object decoders as inductive bias to enable object discovery in a simple auto-encoding setup. Architectures with stronger inductive biases using fixed object size, presence, or propagation priors have been explored in works such as SQAIR~\citep{kosiorek2018sequential} and  SCALOR~\citep{jiang2019scalable}, but generally these methods have faced challenges scaling to more complex real-world data when relying on auto-encoding alone. 
Our work primarily builds on recent advances in object-centric generative models for video sequences~\citep{van2018relational,veerapaneni2020entity,kabra2021simone,kipf2021conditional,zoran2021parts}. Different from our approach, these methods have so far been unable to scale to complex real-world multi-object video data. An alternative class of methods using contrastive learning for object discovery~\citep{kipf2019contrastive,lowe2020learning,xu2022groupvit,henaff2022object}, most notably GroupViT~\citep{xu2022groupvit} and ODIN~\citep{henaff2022object}, has recently achieved some success in discovering semantic groupings in real-world images. However, neither GroupViT nor ODIN model dynamics and typically fail to separate semantically similar object instances in close proximity. In our work, we follow a generative approach, but instead of tasking the decoder to generate complex visual RGB pixel data, we utilize depth information to bootstrap object-centric learning without direct supervision.

\textbf{Object discovery in driving scenes}\quad A range of recent methods~\citep{harley2021track,tian2021unsupervised,zhipeng2022discovering,vobecky2022drivesegment} use a multi-stage pipeline of 1) obtaining \textit{pseudo ground truth} (PGT) segmentation or detection labels via some heuristic, and 2) training a model in a supervised fashion on PGT labels. While this class of methods achieves some success in discovering and tracking objects in real-world driving scenes, it crucially hinges on the quality of the PGT labels, requiring carefully engineered task-specific heuristics to extract objects. Earlier methods solely use clustering heuristics to extract approximate segmentation masks directly from motion trajectories for moving objects~\citep{brox2010object,ochs2011object}. In our work, we instead demonstrate that object segmentation and tracking can emerge in an end-to-end setting on complex real-world data without relying on PGT label generation.

\textbf{Cross-modal learning}\quad For self-supervised object-centric learning from visual data, a range of target modalities and training signals have been explored in the literature. 
By using motion cues from optical flow as prediction targets, several recent methods~\citep{yang2021self,kipf2021conditional} were able to overcome limitations of purely RGB pixel-level generative models, which frequently failed in the presence of complicated textures~\citep{clevrtex2021karazija}. However, this advantage is primarily limited to discovery of moving objects. Utilizing depth targets from a simulator~\citep{bear2020learning} or from sparse LiDAR~\citep{harley2021track,tian2021unsupervised,vobecky2022drivesegment}, has been explored in an effort to overcome these limitations. Different from prior works utilizing multi-stage pipelines and hand-crafted heuristics for extraction of pseudo-labels from LiDAR~\citep{harley2021track,tian2021unsupervised,vobecky2022drivesegment}, we directly utilize the (sparse) depth signal as target and demonstrate that this can enable emergent object segmentation and tracking on real-world driving data without any additional regularizers or pseudo-labeling techniques.

\textbf{Scaling strategies for vision models}\quad It is common practice to scale architectural capacity with dataset complexity and size, while making use of strong data augmentation when addressing various supervised computer vision tasks~\citep{he2016deep,kolesnikov2020big,carion2020end,dosovitskiy2020image}. Nonetheless, self-supervised methods for end-to-end object discovery have primarily been relying on overly simplistic and low-capacity backbone architectures~\citep{greff2019multi,locatello2020object,zoran2021parts}, likely due to the simplicity of datasets and tasks considered in prior work. By scaling object-centric methods to larger, visually more complex datasets, we find that utilizing stronger visual backbone architectures---in combination with data augmentation---can provide substantial benefits. For simplistic datasets with lower visual complexity (and same number of examples), we found anecdotal evidence in preliminary experiments for the opposite effect: both architecture scaling and data augmentation can negatively affect object discovery performance, likely explaining why prior works have not explored these strategies.

\textbf{Depth estimation}\quad Recent advances in supervised monocular depth estimation (see~\citet{ming2021deep} for a review) could be combined with our method in future work, for instance using ordinal regression losses~\cite{fu2018deep}, transformer architectures~\cite{ranftl2021vision}, or more complex instance-wise decoder architectures~\cite{wang2020sdc}.

%% file: methods.tex
\section{Methods}
\label{sec: methods}

\begin{figure}[t!]
    \centering
    \includegraphics[width=\textwidth,trim={0 0.1cm 0 0.1cm},clip]{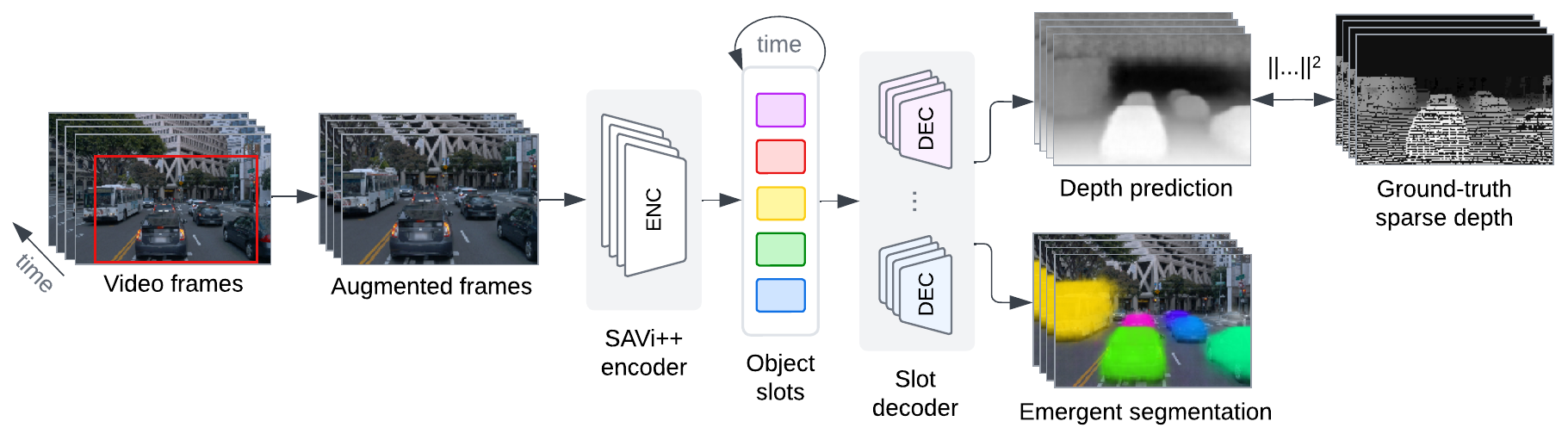}
    \caption{\modelname is an object-centric video model based on \textit{Slot Attention for Video}~\citep{kipf2021conditional}, which encodes a video into a set of temporally-consistent latent variables (\textit{object slots}). Input frames and prediction targets are augmented using random crop augmentations. Augmented frames are passed through the improved SAVi++ encoder and  mapped onto object slots using an attention mechanism~\citep{locatello2020object}. Slots are updated recurrently for each frame and subsequently decoded independently into a depth map and per-slot alpha masks. \modelname is trained using (sparse) depth targets, leading to emergence of temporally-consistent object segmentation in the decoded alpha masks.}
    \label{fig:savi++}
    \vspace{-0.8em}
\end{figure}

We begin by providing a brief introduction to Slot Attention for Video (SAVi), which is the starting point for our exploration. With SAVi++, we introduce several
simple yet crucial improvements, which allow us to bridge the gap to complex real-world data. Our framework is summarized in Figure~\ref{fig:savi++}.

\subsection{Background}
Slot Attention for Video (or SAVi) is a recent state-of-the-art architecture for learning object-centric representations from video with minimal supervision.
We briefly highlight some of its key components below and refer the reader for complete details to~\citet{kipf2021conditional}.

SAVi can be viewed as an autoregressive encoder-decoder video model with a structured latent state composed of $K$ object slots.
At a given time-step, an \emph{encoder} first encodes the observed video frame to yield high-level image features that are useful for learning about objects.
This is followed by Slot Attention \citep{locatello2020object} (the \emph{`corrector'}), which updates the slots using these features and encourages individual slots to specialize to different parts of the observation.
The content of each slot is decoded separately using a \emph{decoder}, which additionally outputs a pixel-level alpha mask to indicate how the decoded values for each slot should be combined.
Together, the mask and decoded slots determine the output of the model at the current time-step from which a loss is computed, e.g., to train the model to predict frame-to-frame motion (optical flow) for this frame.
Slots for the next time-step (for the corrector to update) are obtained by applying a \emph{predictor}, which can model interactions between slots and learn about object dynamics to predict their future state.

In addition to optical flow prediction, SAVi introduces conditioning that helps reduce uncertainty about the part-whole division into objects by pointing the model to specific locations.
Indeed, in the absence of a specific downstream task, scene decomposition can be ambiguous and providing additional information as a conditioning signal may help alleviate this.
The conditioning takes place via the \emph{slot initializer}, which initializes the slots used in the initial video frame.
The initialization may be learned in an unconditional setting (i.e., learn the initial slot states) or obtained by conditioning the initial state on high-level cues such as bounding boxes of objects of interest in the first video frame.
This direction of attention or \emph{input conditioning} 
helped SAVi to succeed in decomposing more complex visual scenes.

\subsection{SAVi++}
As SAVi relies on optical flow prediction as its main training signal for object discovery, its application is primarily limited to settings where all objects in a scene have independent motion. In addition, SAVi struggled to generalize to scenes with a moving camera, even though the optical flow field encodes information about (static) scene geometry in this case.

Here, we identify two key directions for improving SAVi and bridging its capabilities to real-world video data, while preserving its core foundation for learning object representations from video: (1) exploiting depth as a prediction signal, which is readily available in many real-world settings, and (2) utilizing model scaling strategies in terms of encoder improvements and data augmentation, which, despite being commonly used for classic vision problems, are generally underutilized for object-centric learning.
Our improved approach, called \modelname, successfully segments complex objects in naturalistic, real-world video sequences without using direct segmentation or tracking supervision.\looseness=-1

\textbf{Exploiting depth information}\quad
Training object-centric models solely using RGB image or video frame reconstruction proves challenging in the presence of complex visual textures, frequently leading to failure modes such as clustering by color or into object-agnostic spatial regions~\citep{greff2019multi,harley2021track}. In SAVi, optical flow was proposed as a prediction signal to mitigate this issue, while still operating on visual RGB inputs~\citep{kipf2021conditional}.
However, relying solely on optical flow as a prediction target for learning about objects has a clear disadvantage: static objects, which make up the vast majority of visual entities we encounter on a daily basis, are not captured in this signal unless the observer or the entire scene is in motion. As a consequence, SAVi fails to represent objects that are at rest, and similarly struggles with scenes observed from a moving camera, as optical flow can prove challenging to model in this case.

Here, we explore depth as a target signal, used in conjunction with flow or even in isolation. Depth estimation has received little attention in slot-based models, yet does not suffer from the limitation of optical flow in datasets with static objects and camera movement. We thus hypothesize that depth may greatly benefit obtaining emergent object decompositions of complex videos. In terms of practical applicability, we note how depth is a readily-available signal in many real-world settings thanks to the prevalence of RGB-D cameras and LiDAR in settings like self-driving cars~\citep{sun2020scalability}. Even in the absence of depth sensing capabilities, this signal can be cheaply estimated from multi-camera systems~\citep{laga2020survey}.\looseness=-1

In our implementation, we represent the depth signal in image space, which we encode using a log transformation $\log(1 + d)$, where $d$ is the distance of a pixel to the camera (see Figure \ref{fig:movi_samples}). This log-transform puts a stronger emphasis on close-by objects and---in early experiments---we found this form of normalization crucial for reliably training object-centric models using depth targets. \modelname is then trained to minimize the squared difference between the decoder output and this target signal. In case of multiple available targets, such as depth and flow, we concatenate the target images along the channel dimension and predict them using an otherwise unchanged model. 

For sparse targets such as depth obtained from LiDAR, we ignore any points in the image space for which no signal is present in the computation of the loss. 
For LiDAR specifically, we obtain the x, y, z coordinates of all the LiDAR points in the self-driving car (SDC) world and compute the distance of each of the points from the LiDAR sensor. We then use the camera and LiDAR calibration parameters to project the LiDAR point distances from the SDC domain to the camera frame. This projection represents a very sparse approximation of the ground-truth depth signal (Figure \ref{fig:waymo_samples}).

\textbf{Scaling strategies}\quad Visual complexity present in real-world videos necessitates a different class of encoders than those used for simple synthetic datasets. 
Inspired by successful visual backbone architectures for set-based supervised object detection models~\citep{carion2020detr,kamath2021mdetr}, we use a more capable encoder that utilizes the ResNet34~\citep{he2016deep} architecture followed by a transformer encoder~\citep{vaswani2017attention} (with 4 layers, unless otherwise mentioned).
To avoid computation of batch and/or temporal statistics, we replace the typical batch normalization in ResNet34 with group normalization~\citep{wu2018group}. We use a stride 1 convolution and use no max-pooling in the ResNet root block. This results in an overall backbone stride of $8$ (as opposed $32$), which was found to be important for retaining object decomposition capabilities. Please see the appendix for further architectural details.

Drawing inspiration from training schemes commonly used for real-world vision models~\citep{szegedy2015going}, we further apply Inception-style cropping as data augmentation. In particular, we randomly crop a region of each frame with aspect ratio $\in [0.75, 1.33]$ such that enough of the frame is retained after cropping.
The same crop is applied consistently across all frames and the resulting video is resized to the original resolution. Flow fields and depth maps are adjusted accordingly to keep them accurate and spatially aligned with the video frame.

%% file: experiments.tex
\begin{figure}[t]
    \centering
    \begin{subfigure}[b]{0.324\textwidth}
        \centering
        \includegraphics[width=\textwidth]{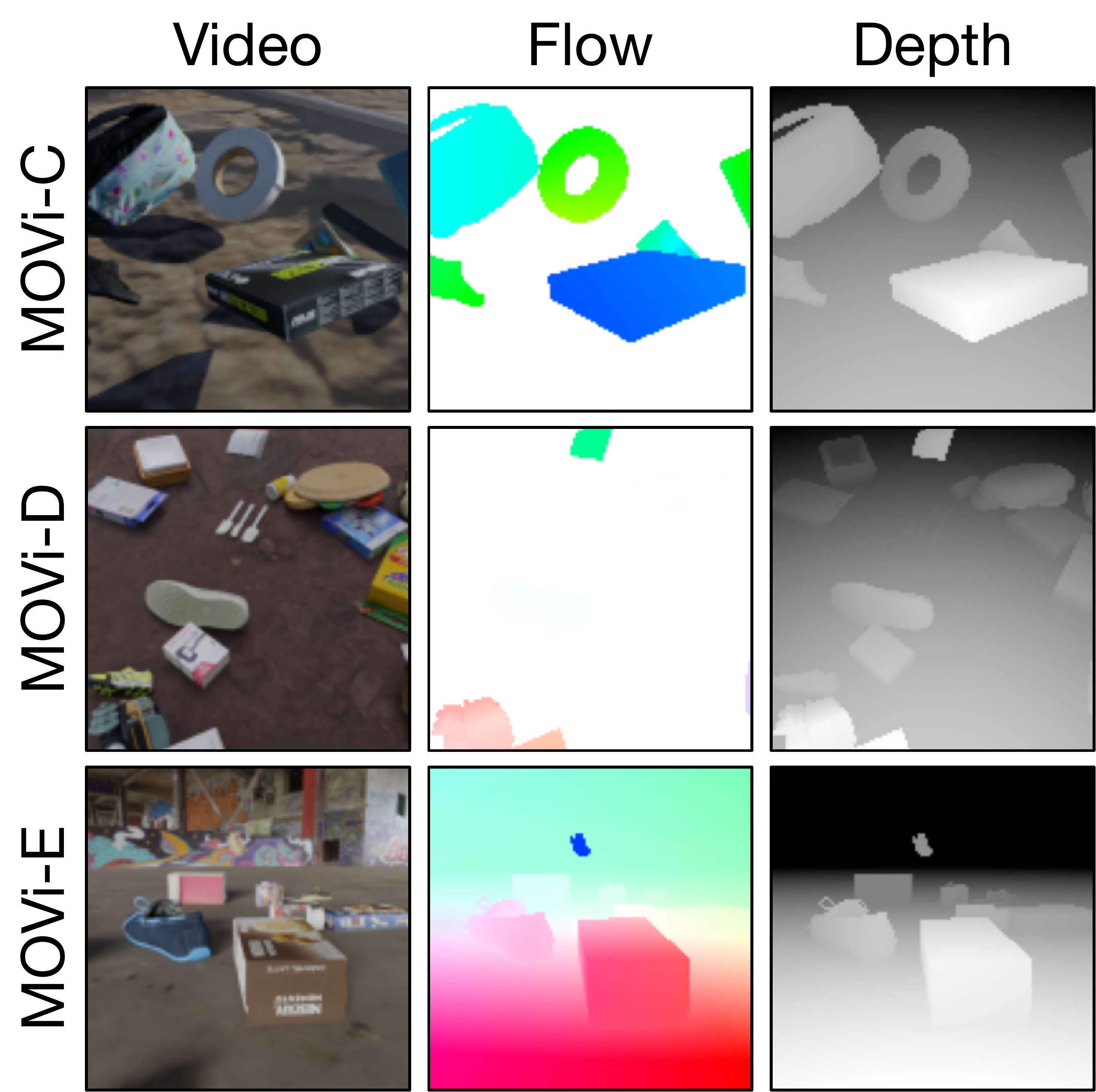}
        \caption{MOVi datasets.}
        \label{fig:movi_samples}
    \end{subfigure}
    \,
    \begin{subfigure}[b]{0.2935\textwidth}
        \centering
        \includegraphics[width=\textwidth]{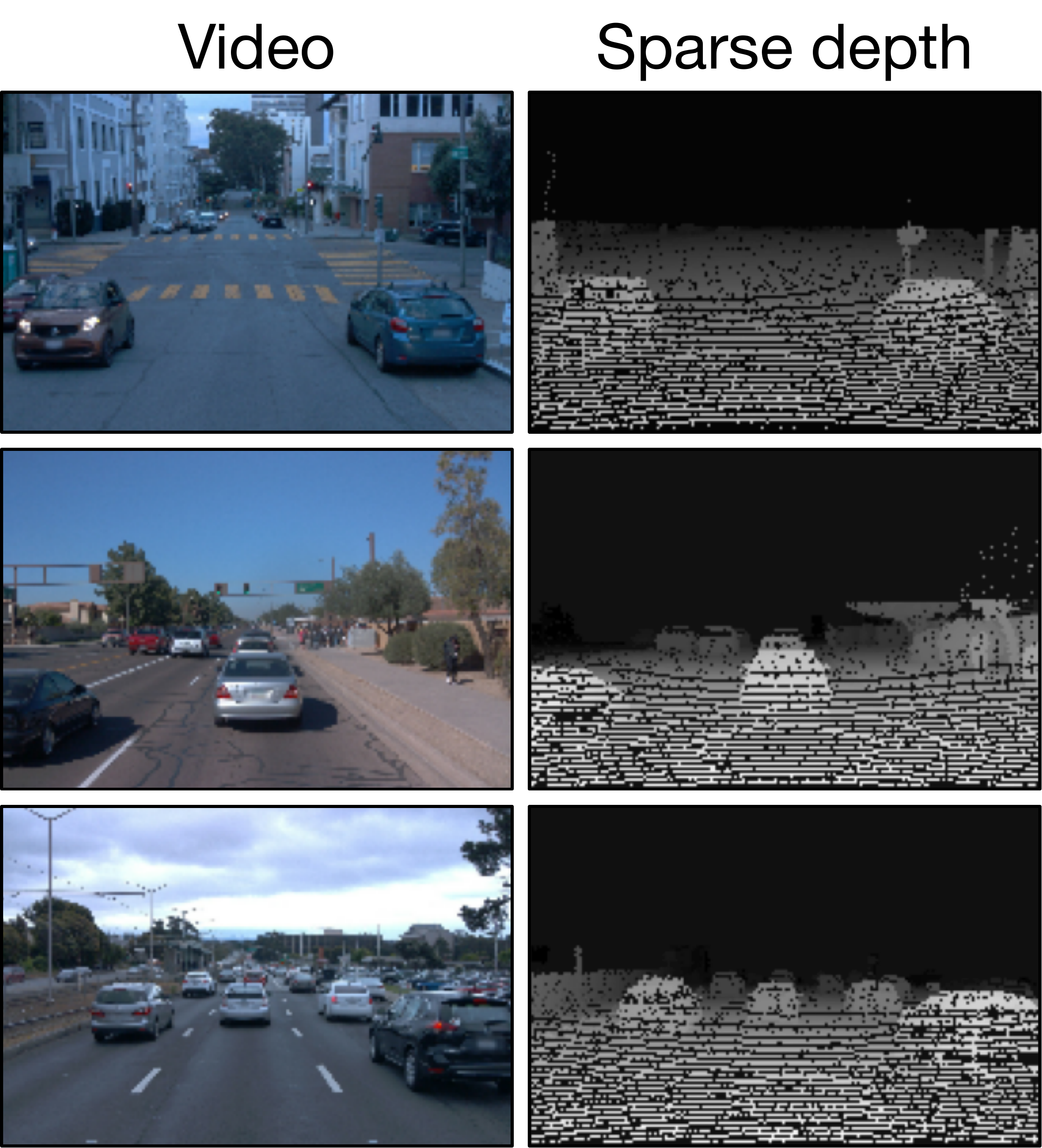}
        \caption{Waymo Open dataset.}
        \label{fig:waymo_samples}
    \end{subfigure}    
    \,
    \begin{subfigure}[b]{0.35\textwidth}
\resizebox{\textwidth}{!}
{
\begin{tabular}{lccccc} \\
 & \rotatebox{90}{Complex texture} & \rotatebox{90}{Moving objects} & \rotatebox{90}{Static objects} & \rotatebox{90}{Moving camera} & \rotatebox{90}{\#frames (train)} \\
\midrule \\[-0.6em]
\makecell{MOVi-C} & \cmark & \cmark & \xmark & \xmark & 234k \\[0.4em]
\makecell{MOVi-D} & \cmark & \cmark & \cmark & \xmark & 234k \\[0.4em]
\makecell{MOVi-E} & \cmark & \cmark & \cmark & \cmark & 234k \\[0.4em]
\midrule 
\makecell{Waymo\\Open} & \cmark & \cmark & \cmark & \cmark & 159.6k
 \end{tabular}
 }
 \caption{Dataset details and statistics.}
 \label{fig:data_stats}
    \end{subfigure}
    \caption{We consider three synthetic Multi-Object Video (MOVi) datasets~\citep{kubric2021} and the large-scale real-world driving dataset Waymo Open~\citep{sun2020scalability}. All datasets contain complex textures and moving objects. The MOVi datasets increase in complexity from MOVi-C (moving objects only) over MOVi-D (+static objects) to MOVi-E (+ moving cameras). Waymo Open contains all these characterisics.}
    \vspace{-1em}
\end{figure}

\section{Experiments}
\label{sec: experiments}
The goal of our experimental evaluation is twofold: 1) on synthetic video data of varying complexity we would like to analyze the potential advantages of utilizing a depth signal and model scaling strategies for learning emergent segmentation and tracking, and 2) we would like to investigate whether these 
improvements enable
bridging the gap to complex real-world video data.

Section~\ref{sec:exp_movi} covers both qualitative and quantitative comparisons of SAVi++ against baselines on the synthetic MOVi datasets. In Section~\ref{sec:exp_ablations}, we perform an ablation study on SAVi++. Finally, in Section~\ref{sec:exp_waymo_open} we demonstrate and analyze results for a SAVi++ model applied to real-world driving videos from the Waymo Open~\citep{sun2020scalability} dataset.

\textbf{Datasets}\quad As basis for our experiments, we use videos of different scene and camera complexities (Figure \ref{fig:data_stats}). We use three synthetic Multi-Object Video (MOVi) datasets (Figure \ref{fig:movi_samples}) introduced in Kubric \citep{kubric2021}, which are created by simulating rigid body dynamics. We narrow our investigation to MOVi datasets with complex naturalistic backgrounds and 3D-scanned everyday objects (variants C, D, and E). MOVi-C is generated using a static camera, and all objects (max.~10) are initialized to move independently. MOVi-D introduces more objects, some of which are dynamic (1-3) and the majority rests statically in the scene (10-20). Finally, MOVi-E introduces random, linear camera movement. Each video contains 24 frames sampled at 12 frames per second (fps).

We also train and evaluate \modelname in a real-world driving setting using the Waymo Open dataset (Figure \ref{fig:waymo_samples}). Waymo Open is comprised of high resolution video data of $1280\times1920$ original resolution from a multi-camera system collected by Waymo vehicles \citep{sun2020scalability}. The dataset consists of 798 train and 202 validation scenes of 20s video each, sampled at 10 fps. We subsample the dataset at 5 fps both for training and validation.  
The dataset also includes LiDAR signals that we use to compute sparse depth maps as discussed in Section~\ref{sec: methods}. 

\textbf{Training setup}\quad For all our experiments, unless stated otherwise, we resize frames to a height of 128 pixels while keeping the aspect ratio fixed, resulting in a $128 \times 128$ resolution for MOVi datasets, and a resolution of $128 \times 192$ for Waymo Open. We train \modelname for 500k steps on Tensor Processing Unit (TPU) accelerators with a batch size of 64 using Adam~\citep{kingma2014adam}.

 We train on randomly sampled sub-sequences of only 6 frames using 24 slots for MOVi and 11 slots for Waymo Open. See appendix for further training details and hyperparameters.

\subsection{\modelname improves object-centric learning on complex synthetic video data}
\label{sec:exp_movi}

We investigate whether the key changes introduced to SAVi~\citep{kipf2021conditional}, which constitute our improved SAVi++ model, allow us to overcome limitations of SAVi and address the most challenging synthetic multi-object video (MOVi) benchmarks introduced in Kubric~\citep{kubric2021}.

\textbf{Setup}\quad We train all models independently on each dataset variant. Both SAVi and SAVi++ are trained in a conditional setting where we initialize slots using ground-truth bounding box information in the first frame. We report the same segmentation metrics as in prior work, i.e.~Foreground Adjusted Rand Index (FG-ARI) ~\citep{rand1971objective,hubert1985comparing} and Mean Intersection over Union (mIoU). FG-ARI is a permutation-invariant clustering similarity metric frequently used for evaluating scene decomposition quality. It compares discovered segmentation masks with ground-truth masks while ignoring any pixels that belong to the background. It is sensitive to temporal consistency of masks, but insensitive to their ordering. The mIoU metric is a standard segmentation metric, here adapted for video as in~\cite{caelles2019the}. We note that this implementation is sensitive to the correct ordering of masks, i.e.~it also measures whether models used the conditioning signal (here, first-frame bounding boxes) correctly.

\textbf{Baselines}\quad Besides comparing to SAVi~\citep{kipf2021conditional}, the most representative prior method for the task we are interested in, we compare against a range of baselines aimed at establishing the difficulty of the unsupervised, bounding box-conditioned video object segmentation task: 1) a bounding box copy (\textit{BBox copy}) baseline, which simply repeats the first-frame boxes throughout the video, 2) a learned \textit{BBox propagation baseline} that does not receive visual inputs, to test for easily exploitable biases in the datasets, 3) k-Means clustering baselines, that cluster the flow and/or depth signal across the video sequence (initialized using the ground-truth object centers in the first frame), and 4) a label propagation baseline, that uses visual features to propagate the initial boxes (rendered as rectangular masks) across the video, based on Contrastive Random Walks (CRW)~\citep{jabri2020space}. See appendix for further details.\looseness=-1

\begin{figure}[t!]

\begin{subfigure}[b]{0.36\textwidth}
    \centering
    \includegraphics[width=\textwidth]{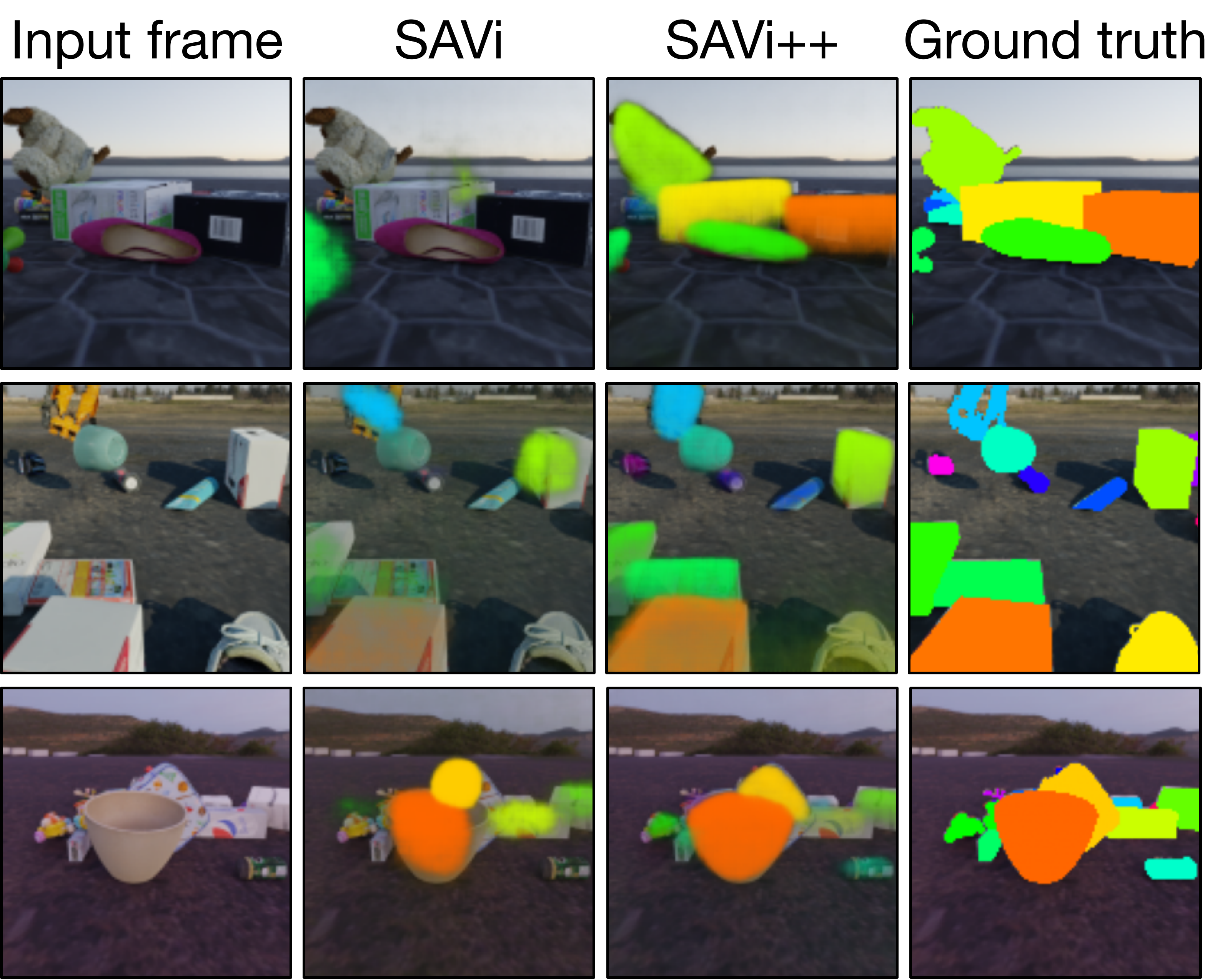}
    \caption{MOVi-E qualitative results.}
    \label{fig:movi_qualitative} 
\end{subfigure}
\quad
\begin{subfigure}[b]{0.6\textwidth}
    \centering
    \includegraphics[width=\textwidth,trim={0 0.1cm 0 0},clip]{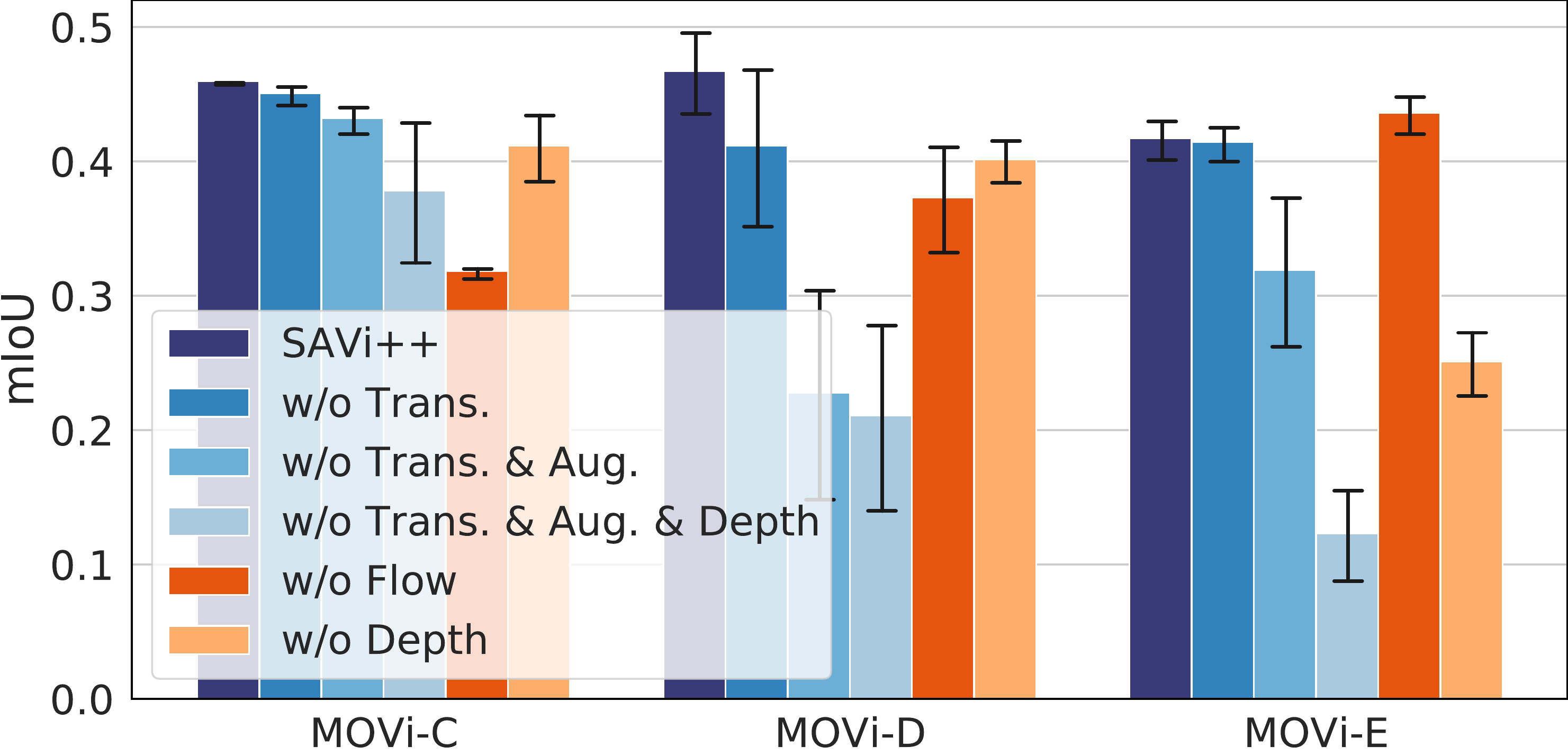}
    \caption{Ablation study}
    \label{fig:ablations}
\end{subfigure}
\caption{\textbf{Left}:  Qualitative results of SAVi++ compared to SAVi~\citep{kipf2021conditional} on the synthetic MOVi-E dataset with camera motion. \textbf{Right}: SAVi++ ablation study on MOVi-C, D, and E. Bars reflect validation set mIoU (mean $\pm$ standard error for 5 seeds). We ablate: 1) the transformer encoder (\textit{w/o Trans.}), 2) data augmentation (\textit{w/o Trans. \& Aug.}), and 3) depth targets (\textit{w/o Trans. \& Aug. \& Depth}). We further report results for training without flow, while only using depth targets (\textit{w/o Flow}).}
\vspace{-0.8em}
\end{figure}

\textbf{Results}\quad
Quantitative results can be seen in Table~\ref{table:main} and qualitative results on MOVi-E in Figure~\ref{fig:movi_qualitative}. 
The \textit{BBox copy} method serves as a trivial baseline which a learning-based approach should outperform.
While the original SAVi model does so on MOVi-C, it clearly fails to model the more complex MOVi-D and -E datasets. The BBox copy baselines is---perhaps unsurprisingly---strongest on MOVi-D, where most objects are static. SAVi++ outperforms this baseline on all datasets, indicating that it learns non-trivial segmentation and tracking capabilities. Indeed, this advantage does not solely come from fitting certain biases in the datasets, as a learned BBox propagation baseline (using the same predictor as in SAVi++) that does not receive visual input, fails to generalize to unseen evaluation videos.
It is worth noting that neither of the MOVi tasks can easily be solved by simply clustering the target signals, as the results for the k-Means baselines demonstrate. 

Compared to CRW it can be seen how \modelname yields markedly better mIoU on MOVi-C and D, while performance on MOVi-E is similar. 
Note that, unlike \modelname, CRW is merely capable of propagating pixel-level annotations across frames in a video and does not by itself produce instance-level object segmentations or corresponding object-representations that could be used for down-stream tasks.
Finally, comparing SAVi++ and SAVi directly, we see that SAVi++ overcomes the primary limitations of SAVi on the harder MOVi-D and -E datasets, both quantatively (Table~\ref{table:main}) and qualitatively (Figure~\ref{fig:movi_qualitative}), while also improving performance on MOVi-C.

\textbf{Discussion}\quad 
It is evident that a small number of critical changes to SAVi~\citep{kipf2021conditional}, namely utilizing depth targets, a stronger architecture, and data augmentation, can have dramatic consequences on the ability of this slot-based model to learn emergent object segmentation and tracking in complex video sequences. The difference between \modelname and SAVi is especially evident for the more complex datasets in our study (e.g., improving the mIoU score on MOVi-E from 30.7\% to 47.1\%; see also Figure \ref{fig:movi_qualitative}). These results demonstrate that \modelname is better suited for various data complexities in terms of object dynamics and camera movement, which are likely to exist in real-world data.

\begin{table}[t]
\caption{MOVi results in terms of mean score $\pm$ standard error (5 seeds) from evaluating SAVi++ and baseline models on validation set video sequences of increased length (24 frames). *: we use the official implementation of CRW~\citep{jabri2020space}, which does not report FG-ARI.}
\vspace{-0.5em}
\label{table:main}
\centering
\resizebox{\textwidth}{!}
{
\begin{tabular}{lcccccc} \\
\toprule
 & \multicolumn{3}{c}{\textbf{mIoU$\uparrow$} (\%)}  & \multicolumn{3}{c}{\textbf{FG-ARI$\uparrow$} (\%)} \\
 \cmidrule(l{2pt}r{2pt}){2-4} \cmidrule(l{2pt}r{2pt}){5-7}
\textbf{Model} & \textbf{MOVi-C} & \textbf{MOVi-D} & \textbf{MOVi-E} & \textbf{MOVi-C} & \textbf{MOVi-D} & \textbf{MOVi-E} \\
\cmidrule(l{2pt}r{2pt}){1-4} \cmidrule(l{2pt}r{2pt}){5-7}
BBox copy & $12.3\hphantom{\pm0.0}$ &  $42.8\hphantom{\pm0.0}$  & $32.9\hphantom{\pm0.0}$ & $11.8\hphantom{\pm0.0}$ & $68.0\hphantom{\pm0.0}$ & $54.7\hphantom{\pm0.0}$ \\
BBox propagation &
$22.9 \scriptstyle{\,\pm\,0.1}$ & 
$26.7 \scriptstyle{\,\pm\,0.8}$	 & 
$24.1 \scriptstyle{\,\pm\,1.1}$ &
$9.6 \scriptstyle{\,\pm\,0.5}$ & 
$24.9 \scriptstyle{\,\pm\,3.7}$	 &
$18.4 \scriptstyle{\,\pm\,3.9}$
 \\
K-Means (depth)     & 
$\hphantom{0}7.1\scriptstyle{\,\pm\,0.3}$ &
$\hphantom{0}6.0\scriptstyle{\,\pm\,0.4}$ &
$\hphantom{0}5.4\scriptstyle{\,\pm\,0.3}$ &
$26.3\scriptstyle{\,\pm\,1.0}$ &
$30.9\scriptstyle{\,\pm\,0.7}$ &
$32.2\scriptstyle{\,\pm\,0.6}$  \\
K-Means (flow)     & 
$10.7\scriptstyle{\,\pm\,0.5}$ &
$\hphantom{0}7.4\scriptstyle{\,\pm\,0.4}$ &
$\hphantom{0}6.0\scriptstyle{\,\pm\,0.3}$ & 
$26.5\scriptstyle{\,\pm\,1.0}$ &
$30.9\scriptstyle{\,\pm\,0.8}$ &
$33.1\scriptstyle{\,\pm\,0.7}$  \\
K-Means (flow+depth)     & 
$10.6\scriptstyle{\,\pm\,0.6}$ &
$\hphantom{0}6.7\scriptstyle{\,\pm\,0.4}$ &
$\hphantom{0}5.3\scriptstyle{\,\pm\,0.3}$ &
$26.6\scriptstyle{\,\pm\,1.0}$ &
$35.9\scriptstyle{\,\pm\,1.0}$ &
 $34.8\scriptstyle{\,\pm\,0.7}$ \\
CRW~\citep{jabri2020space} &
$27.8\scriptstyle{\,\pm\,0.2}$ &
$45.3\scriptstyle{\,\pm\,0.0}$&
$\mathbf{47.5}\scriptstyle{\,\pm\,0.1}$ & * & * & * \\
SAVi~\citep{kipf2021conditional} & 
$43.1 \scriptstyle{\,\pm\,0.7}$	 & 
$22.7 \scriptstyle{\,\pm\,7.5}$ & 
$30.7 \scriptstyle{\,\pm\,4.9}$ &
$77.6 \scriptstyle{\,\pm\,0.7}$ & 
$59.6 \scriptstyle{\,\pm\,6.7}$ & 
$55.3 \scriptstyle{\,\pm\,5.8}$ 
 \\

\modelname (ours) & 
$\mathbf{45.2} \scriptstyle{\,\pm\,0.1}$ 		 
& 
$\mathbf{48.3} \scriptstyle{\,\pm\,0.5}$
& 
$\mathbf{47.1} \scriptstyle{\,\pm\,1.3}$
&
$\mathbf{81.9} \scriptstyle{\,\pm\,0.2}$		&
$\mathbf{86.0} \scriptstyle{\,\pm\,0.3}$			&
$\mathbf{84.1} \scriptstyle{\,\pm\,0.9}$
\\ 
\bottomrule
\end{tabular}
}
\end{table}

\subsection{Ablation study}
\label{sec:exp_ablations}

In this section, we report results of an ablation study to gauge the contribution of the different components of \modelname. The three main ingredients of \modelname are 1) the use of depth as training target, 2) the extra capacity added to SAVi by including a transformer encoder, and 3) the use of data augmentation. Figure \ref{fig:ablations} shows a systematic ablation of each of those components. Removing the transformer encoder reduces object segmentation quality, yet the degradation in performance is relatively limited. While data augmentation only has a mild effect on the simpler MOVi-C dataset, it makes a substantial difference on the more challenging datasets, MOVi-D and E. Finally, removing depth targets reduces performance further and is particularly catastrophic on MOVi-E.

In fact, we find that training \textit{solely} using depth targets without relying on predicting optical flow as well (see \textit{w/o Flow} in Figure~\ref{fig:ablations}) still allows the model to accurately segment and track objects, especially on the more complex MOVi-D and E datasets. This result is particularly strong on MOVi-E where jointly predicting optical flow presents a difficult task for scenes with camera movement. Further, training on depth targets was very crucial to obtain good performance on the most complex synthetic data MOVi-E as demonstrated with the large drop in mIoU when ablating depth and relying only on optical flow to train the model.

\begin{figure}[t!]
    \centering
    \includegraphics[width=0.9\textwidth]{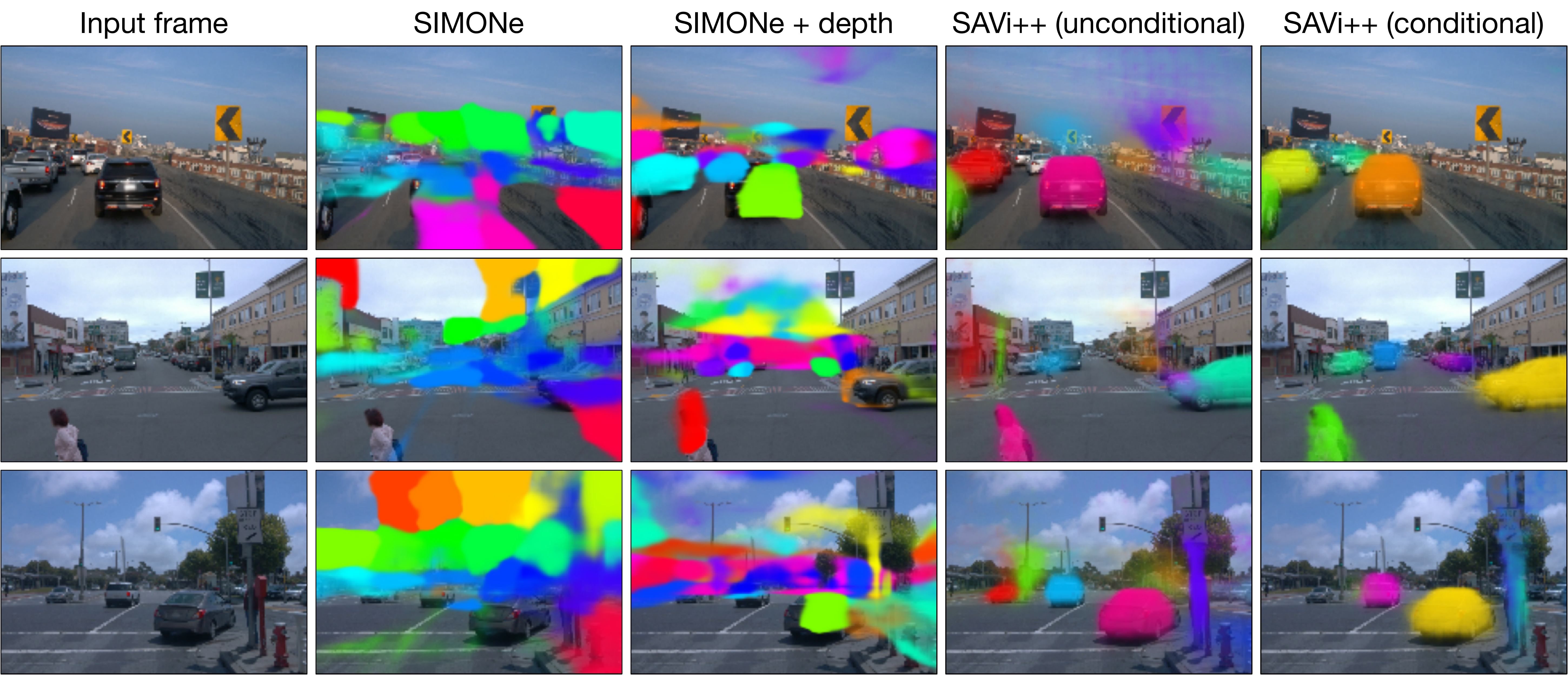}
    \caption{Qualitative segmentation comparison on the Waymo Open validation set. Naive application of a SIMONe~\citep{kabra2021simone} baseline model to this dataset results in failure, while adapting SIMONe to predict (sparse) depth maps yields rough (but frequently misaligned) segmentation masks. SAVi++ generally produces highly accurate segmentation masks, while its unconditional results are promising. Here, we hide masks that occupy more than 1300 pixels on average per frame to ease interpretability.}
    \label{fig:waymo_qualitative}
    \vspace{-0.8em}
\end{figure}

\subsection{\modelname enables emergent segmentation on real-world driving data}
\label{sec:exp_waymo_open}

In the previous section, we found that solely using depth as a training target can be sufficient to learn emergent object segmentation and tracking.  This finding provides a strong motivation for scaling this class of methods to real-world data, where the availability of optical flow relies on approximate and potentially inaccurate flow estimation methods, whereas depth can be accurately measured using technologies like LiDAR. To investigate this possibility, we use the Waymo Open dataset~\citep{sun2020scalability}, which includes videos obtained from cameras mounted on cars in various traffic environments.

\textbf{Setup}\quad To obtain a depth signal, we project 3D LiDAR points into the camera frame, resulting in a very sparse depth image for each time step (see Figure \ref{fig:waymo_samples} for examples). We exclude pixels that do not have a valid LiDAR point when computing the L2 loss in image space.
We train \modelname with 11 slots on 6 frames and evaluate the model on sequences of 10 frames. Due to the absence of ground-truth segmentation labels in Waymo Open, we quantitatively measure performance compared to ground-truth bounding boxes using three metrics. The Center-of-Mass (CoM) distance measures the average Euclidean distance between the centroid of the predicted segmentation masks and the centers of the ground-truth bounding boxes. We report the centroid distance normalized by the maximum achievable distance in the video frame.
Additionally, we separately measure the fraction of cases where any sort of segment is predicted when a valid ground-truth box exists, denoted as bounding box recall (B.~Recall).
The Bounding Box mIoU (B.~mIoU) is analog to mIoU using predicted and ground-truth bounding boxes. The former are obtained by training a readout MLP to predict bounding boxes from the slot representations.
See appendix for further details.

\textbf{Baselines}\quad We quantitatively compare to the subset of previous baselines that work with (sparse) depth.
Further, we report qualitative results for SAVi++ in the unconditional setting, i.e.~without providing first-frame bounding boxes to the model to initialize slots, and compare to SIMONe~\citep{kabra2021simone} as a representative object-centric video model baseline from the literature.

\begin{figure}[t!]
    \centering
    \includegraphics[width=\textwidth]{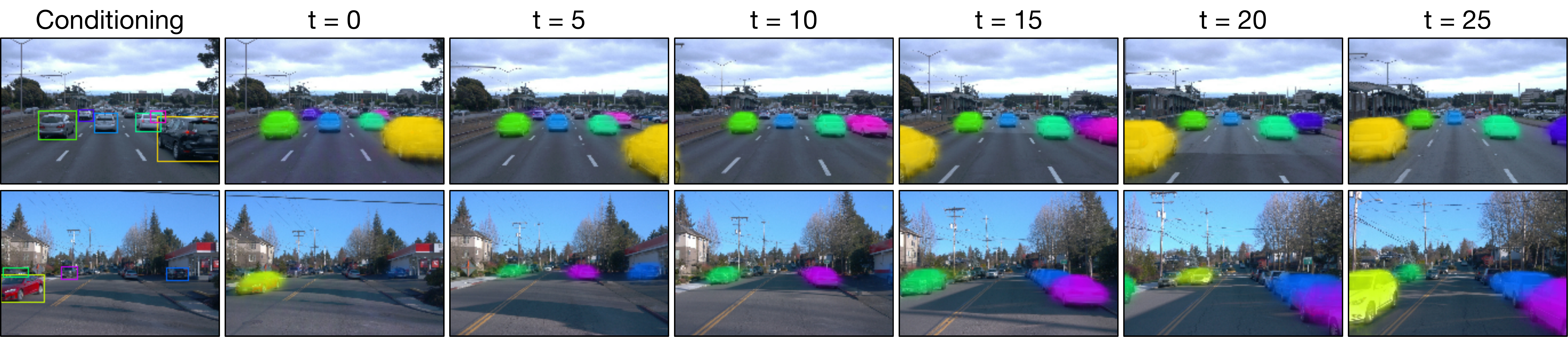}
    \caption{Waymo Open qualitative results of SAVi++ (conditional) over long sequences.} 
    \label{fig:waymo_long}
    \vspace{-1em}
\end{figure}

\textbf{Results}\quad
Quantitative results can be seen in Table~\ref{table: waymo conditional} and qualitative results in Figures~\ref{fig:waymo_qualitative}--\ref{fig:waymo_long}.
We find that \modelname markedly outperforms the BBox copy and propagation baselines, as well as the clustering baseline in terms of object tracking.
Further, the bounding box recall is high indicating that valid objects are rarely ignored.
The qualitative results in Figures~\ref{fig:waymo_qualitative}--\ref{fig:waymo_long} even better reflect the significance of \modelname's performance as well as its potential utility for object-centric representation learning from real-world videos (for \modelname results divided per object category see Table \ref{table:waymo:percls}).

Our results using sparse depth targets suggest that \modelname does not need complete (i.e.~dense) depth supervision. To investigate how \textit{accurate} this signal needs to be, we explored the degree of sensitivity of \modelname to noise in the depth signal. We trained \modelname with noisy depth targets by applying additive Gaussian noise to the ground-truth sparse LiDAR depth signals with standard deviations of 10cm, 20cm and 40cm. We found that \modelname was able to retain its emergent tracking performance even at the highest considered noise scale of 40cm (see Table \ref{table: noisy waymo} in Appendix).

\begin{wraptable}{r}{0.5\textwidth}
\vspace{-1em}
\caption{Waymo Open results (mean $\pm$ standard error in \%, 3 seeds) from evaluating models on sequences of 10 frames. \modelname HR is a variant trained on higher-resolution ($256\times384$) video frames.}
\label{table: waymo}
\label{table: waymo conditional}
\vspace{-0.8em}
\centering
\resizebox{\linewidth}{!}{
\begin{tabular}{lccc} \\
\toprule 
& & (\%) \\
\textbf{Model} & \textbf{CoM$\downarrow$}  & \textbf{B.~mIoU$\uparrow$} & \textbf{B.~Recall $\uparrow$}\\
\midrule 
BBox Copy & 
$5.0$
& $44.3$ & 100\\
BBox Prop. & 
$5.1 \scriptstyle{\,\pm\,0.1}$ 
& $38.5 \scriptstyle{\,\pm\,0.5}$ & 100\\
K-Means (depth)  & 
$13.0 \scriptstyle{\,\pm\,0.1}$ & 
-- & 100\\
SAVi (RGB) & 
$21.5 \scriptstyle{\,\pm\,1.8}$ &
$7.9 \scriptstyle{\,\pm\,0.9}$ & $95.8 \scriptstyle{\,\pm\,2.7}$\\ 
SAVi (depth) & 
$24.7 \scriptstyle{\,\pm\,0.7}$ &
$10.3 \scriptstyle{\,\pm\,2.4}$ &  
$97.4 \scriptstyle{\,\pm\,0.6}$
\\
\modelname        & 
$\mathbf{4.4} \scriptstyle{\,\pm\,0.2}$
&
$\mathbf{49.7} \scriptstyle{\,\pm\,0.7}$ & 
$96.5 \scriptstyle{\,\pm\,0.7}$
\\
\midrule 

\modelname  HR      & 
$\mathbf{3.9} \scriptstyle{\,\pm\,0.1}$  
&
$\mathbf{51.9} \scriptstyle{\,\pm\,0.4}$ & 
$96.2 \scriptstyle{\,\pm\,0.4}$
\\

\midrule 

Supervised  & 
$1.1 \scriptstyle{\,\pm\,0.0}$
&
$67.6 \scriptstyle{\,\pm\,0.6}$ & \\
\bottomrule
\end{tabular}
}
\vspace{-0.8em}
\end{wraptable}

We additionally experimented with removing the bounding box conditioning in \modelname in the initial frame.
Removing this conditioning signal and using a learned initialization together with a simplified encoder also yielded good object decompositions (see \modelname (unconditional) in Figure \ref{fig:waymo_qualitative}).
Compared to using plain SIMONe~\citep{kabra2021simone}, we observe that \modelname (unconditional) performs markedly better.
Interestingly, modifying the non-autoregressive SIMONe baseline similar to \modelname by predicting sparse depth instead of RGB also showed improvement in object emergence.
This gives further evidence that using depth is suitable for learning object-centric representations from real videos. Quantitatively, SAVi++ achieves a CoM distance of $6.9\pm0.5$ while SIMONe (with depth loss) achieves $7.4\pm0.2$\footnote{These baseline results are improved compared to an earlier version of the paper by using exactly the same depth target transformation as for SAVi++.} over a sequence of 12 frames at test time, evaluated using Hungarian matching.\looseness=-1

We show qualitative results for longer sequences in Figure~\ref{fig:waymo_long} and in video-format in the supplementary material. It is worth noting that \modelname was only trained on 6 frames and did not receive any tracking supervision. Interestingly, we find that objects are often consistently tracked until the moment they leave the scene. At this stage, slots are freed up again and tend to bind to previously unexplained or new objects. This behaviour indicates that our reported tracking metrics are an underestimation of the capabilities of the model, as such re-binding is not accounted for.
It is, however, conceivable that re-binding events could be identified post-hoc if one were to use the representations learned by \modelname for downstream tasks, which is an interesting avenue for future work.

\subsection{Limitations}

With \modelname, we demonstrated the first proof of concept that an emergent object-centric decomposition of real-world complex videos is possible with an end-to-end slot-based approach. Yet, there is still a lot of room for improvement.

\textbf{Reliance on conditioning}\quad We focused our exploration on the conditional setup where we provided cues in the form of bounding boxes of objects in the first frame. Although the use of such ``object hints'' may share some similarity to how human visual attention (and how humans parse a visual scene) can be directed via external signals (e.g., via gestures such as pointing), it ultimately limits the practical applicability of our approach.
Preliminary results with unconditional \modelname suggest that this information may not be strictly necessary and could be removed in future research.

\textbf{Reliance on ground-truth target signals}\quad 
In a similar vein, the reliance of \modelname on ground-truth target signals for training is a limitation that may affect its practical applicability.
Fortunately, LiDAR sensors for depth estimation are readily available in many application domains (such as in robotics and self-driving), and there is also a rich literature on monocular depth estimation.
While estimated, depth (or flow) are expected to be noisier compared to the signals considered in our experiments, our experiment with ``noisy depth'' offers an initial sign that this may not affect performance much.

\textbf{Gap to videos recorded in the wild}\quad
It is also important to point out that although Waymo Open offers a challenging real-world benchmark for learning about objects, its videos are relatively structured compared to real-world videos recorded ``in the wild'', and especially heavy on cars, roads, traffic signs, pedestrians, etc.
Other datasets, such as DAVIS~\citep{perazzi2016benchmark} or Kinetics~\citep{kay2017kinetics} offer greater complexity in that regard and it is foreseeable that further development of \modelname will be needed to truly support these.
An example of this is that objects in Waymo Open usually do not re-appear, which is an aspect that is currently not explicitly modeled in \modelname (e.g.~to ensure that the same object is re-captured by the same slot). 
More generally, there is substantial headroom to improve the modeling of disappearing and reappearing objects in future work, such as by explicitly modeling object presence \citep{kosiorek2018sequential}, or by explicitly attending to past latent states \citep{zhou2022slot}.

\textbf{Gap to supervised approaches}\quad
Finally, we note how both in the conditional and the unconditional setting, the segmentation and tracking performance, though impressive given the minimal amount of supervision the model receives, still qualitatively lags behind supervised approaches. Improving on the temporal consistency of object tracks, especially in the unconditional setting, is another promising direction for future work.

%% file: conclusion.tex
\section{Conclusion}
\label{sec: conclusion}

We demonstrate that object tracking and segmentation can emerge from utilizing information about scene geometry in the form of depth signals in complex video data with slot-based neural architectures.
We utilize a series of synthetic multi-object video benchmarks with increasing complexity to find a simple yet effective set of changes to an existing state-of-the-art object-centric video model (SAVi), allowing us to bridge the gap from synthetic to complex real-world driving videos.

Our work marks a first step towards building end-to-end trainable systems that learn to perceive the world in an object-centric, decomposed fashion without relying on detailed human supervision. While many open challenges remain, this result evidences that object-centric deep neural networks are not inherently limited to simple synthetic environments, and we are excited about the potential for this class of methods to radically reduce the need for human supervision in building scalable perceptual systems for the real world.

%% file: appendix.tex
\appendix

\section{Societal impact}
\label{sec: societal impact}
Our work focuses on extending object-centric representation learning to real-world videos. This class of methods has the potential for enabling systems to more reliably solve down-stream tasks requiring object representations, such as relational reasoning over entities in videos, and providing better interpretation of model decisions. Some applications that may benefit from this approach include perception in autonomous vehicles, robotics and classic computer vision problems such as object detection. While the method demonstrated in this paper is still far from a state where it could be directly employed in computer vision applications, we would like to raise awareness that---as with most methods developed for computer vision---advances in this field might also aid the development applications with potential negative societal impact such as surveillance.

\section{Additional results}
\label{sec:additional_results}

\begin{figure}[htp]
    \centering
    \includegraphics[width=0.9\textwidth]{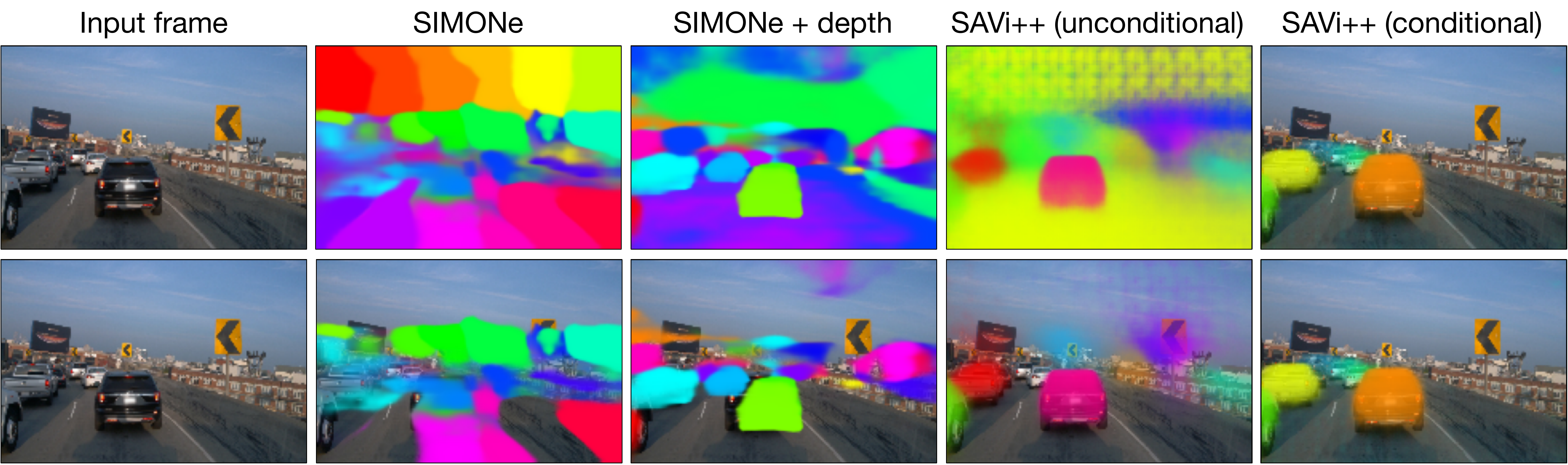}
    \caption{Comparison of qualitative result visualization without (top) and with (bottom) mask thresholding. We apply a simple thresholding heuristic of dropping any masks that (on average across frames) occupy more than 1300px per frame, which aids interpretability. Thresholding does not have an effect on the conditional model.} 
    \label{fig:waymo_threshold} 
    \vspace{-1em}
\end{figure}

\begin{figure}[htp]
    \centering
    \includegraphics[width=\textwidth]{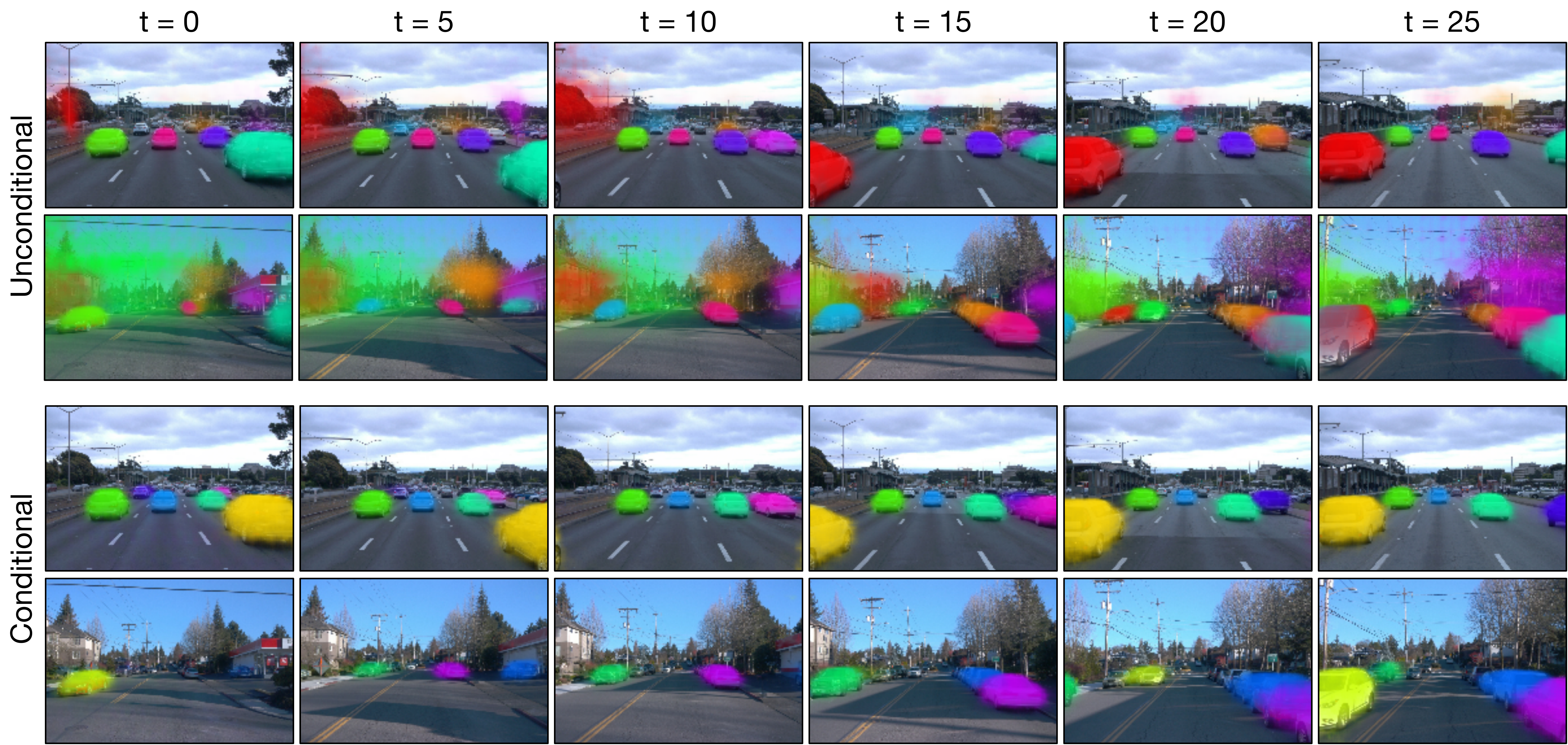}
    \caption{Qualitative results on long Waymo Open validation set videos (for SAVi++ models trained on 6 frames). Top: Results for an unconditional SAVi++ model, trained and evaluated without bounding box conditioning in the first frame. Bottom: Conditional setting shown for reference.} 
    \label{fig:waymo_long_unconditional} 
    \vspace{-1em}
\end{figure}

\paragraph{Qualitative results} In Figure~\ref{fig:waymo_threshold}, we show the effect of our mask thresholding heuristic applied for our unsupervised model visualizations. The intention for this simple heuristic is to aid interpretability of the discovered object segmentation masks. We further show qualitative results for emergent tracking on long sequences in the unconditional setting (i.e.~without bounding box conditioning) in Figure~\ref{fig:waymo_long_unconditional}. Compared to the conditional setting (shown for reference), tracking is less consistent and slots explain not only cars, but also environmental objects or part of the background.

\paragraph{Quantitative results} 
We additionally compare SAVi and SAVi++ on the synthetic MOVi-A and MOVi-B datasets~\citep{kubric2021}.
In MOVi-A, scenes consist of a gray floor, four light sources, a fixed static camera and between 3 and 10 simple geometric objects that vary in terms of their shape (cube, sphere, cylinder), material (rubber, metal), size (small, large), and color (blue, brown, cyan, gray, green, purple, red, yellow).
MOVi-B is a straightforward extension, which adds additional variation to the object shape, size, and color; background color; and static camera positions.

In Table~\ref{table:movi-A/B} it can be seen how \modelname performs similar or worse in terms of mIoU and FG-ARI on these datasets.
We mainly attribute this to our strategy for scaling in \modelname, which appears susceptible to overfitting on these much simpler domains.
Additionally, the benefit of using depth information is limited, since all objects are in motion for these datasets.
The contrast between the results presented in the main paper for MOVi-C to -E and Table~\ref{table:movi-A/B} (MOVi-A and -B) emphasizes the importance of considering benchmarks that are more representative of the real world for model development.

\begin{table}[!ht]
\caption{MOVi results in terms of mean score $\pm$ standard error (5 seeds) from evaluating \modelname and SAVi models on validation set video sequences of increased length (24 frames).}
\vspace{-0.5em}
\label{table:movi-A/B}
\centering
\begin{tabular}{lcccc} \\%  cc}
\toprule
 & \multicolumn{2}{c}{\textbf{mIoU$\uparrow$} (\%)}  & \multicolumn{2}{c}{\textbf{FG-ARI$\uparrow$} (\%)} \\
 \cmidrule(l{2pt}r{2pt}){2-3} \cmidrule(l{2pt}r{2pt}){4-5}
\textbf{Model} & \textbf{MOVi-A} & \textbf{MOVi-B} & \textbf{MOVi-A} & \textbf{MOVi-B} \\
\cmidrule(l{2pt}r{2pt}){1-3} \cmidrule(l{2pt}r{2pt}){4-5}
SAVi~\citep{kipf2021conditional} & 
$82.3 \scriptstyle{\,\pm\,0.3}$	& 
$44.5 \scriptstyle{\,\pm\,9.3}$ & 
$96.8 \scriptstyle{\,\pm\,0.4}$ & 
$73.9 \scriptstyle{\,\pm\,10.7}$ 
 \\
\modelname & 
$76.1 \scriptstyle{\,\pm\,0.9}$	& 
$25.8 \scriptstyle{\,\pm\,11.3}$ &
$98.2 \scriptstyle{\,\pm\,0.2}$	&
$48.3 \scriptstyle{\,\pm\,15.7}$		
\\
\bottomrule
\end{tabular}
\end{table}

\begin{table}[h!]
\caption{Breakdown of \modelname results from Table~\ref{table: waymo conditional} in terms of three classes of objects: car, person, and cyclist.}
\label{table:waymo:percls}
\vspace{-0.8em}
\centering
\begin{tabular}{lccc} \\
\toprule 
Metric & Car & Person & Cyclist\\
\midrule 
Num. objects &  $15350$ & $2102$ & $275$ \\
\textbf{CoM}$\downarrow$ & \ebar{4.2}{0.2} & \ebar{6.4}{0.1} & \ebar{1.9}{0.1} \\
\textbf{B. mIoU}$\uparrow$ & \ebar{52.5}{0.8} & \ebar{27.0}{0.5} & \ebar{42.2}{2.1} \\
\textbf{B. Recall}$\uparrow$ & \ebar{96.7}{0.7} & \ebar{95.1}{1.1} & \ebar{99.3}{0.3} \\
\bottomrule
\end{tabular}
\end{table}

We report per category results for \modelname on the WaymoOpen dataset in Table~\ref{table:waymo:percls}. We find that, as expected, this dataset is dominated by cars and performance is very good in this category. \modelname also performs very well on cyclists, which are very rare in this dataset, and indicates that \modelname has not overfit to blobby car like objects.

We further report results for SAVi++ under the influence of noise in the sparse depth targets on Waymo Open in Table~\ref{table: noisy waymo}. Our results indicate that emergent tracking performance is largely unaffected by noise scales (standard deviations) of up to $\sigma=40\textnormal{ cm}$.

\begin{table}
\caption{Waymo Open results (mean $\pm$ standard error in \%, 3 seeds) from evaluating models on sequences of 10 frames. \modelname was trained with noisy depth targets with standard deviation ($\sigma$) specified below.}
\label{table: noisy waymo}
\vspace{-0.8em}
\centering
\begin{tabular}{lccc} \\
\toprule 
& & (\%) \\
\textbf{Model} & \textbf{CoM$\downarrow$}  & \textbf{B.~mIoU$\uparrow$} & \textbf{B.~Recall $\uparrow$}\\
\midrule 
\modelname        & 
${4.4} \scriptstyle{\,\pm\,0.2}$ &
${49.7} \scriptstyle{\,\pm\,0.7}$ & $96.5 \scriptstyle{\,\pm\,0.7}$
\\
\modelname ($\sigma = 10\textnormal{ cm}$) & 
$4.3 \scriptstyle{\,\pm\,0.2}$ & 
$50.1 \scriptstyle{\,\pm\,0.3}$ & 
$96.8 \scriptstyle{\,\pm\,0.3}$\\
\modelname ($\sigma = 20\textnormal{ cm}$) & 
$4.3 \scriptstyle{\,\pm\,0.2}$ & 
$49.7 \scriptstyle{\,\pm\,0.3}$ & 
$97.4 \scriptstyle{\,\pm\,0.5}$\\
\modelname ($\sigma = 40 \textnormal{ cm}$) & 
$4.2 \scriptstyle{\,\pm\,0.0}$ & 
$50.1 \scriptstyle{\,\pm\,0.3}$ & 
$96.9 \scriptstyle{\,\pm\,0.5}$\\
\bottomrule
\end{tabular}
\end{table}

\paragraph{Supplementary videos} We provide several video results in the supplementary material for both SAVi++ (conditional, uncondtional, and high-resolution model variants) and the SIMONe~\citep{kabra2021simone} baseline (both in its original form and in our adapted depth-prediction variant).

\section{Training setup}

We train our models for 500k steps (300k steps for the ablation study) on Tensor Processing Unit (TPU) accelerators with a batch size of 64 using Adam~\citep{kingma2014adam}. We linearly increase the learning rate for 2500 steps to $0.0002$ (starting from 0) and then decay the learning rate with a Cosine schedule~\citep{loshchilov2016sgdr} back to 0 for the rest of the training steps. We clip the gradients to a global norm value of 0.05 to stabilize training. To create video examples for training, we split each video into sub-sequences of 6 frames each. We use a total of 24 slots on MOVi and 11 slots on Waymo Open for \modelname models. We use 1 iteration per frame for the Slot Attention~\citep{locatello2020object} module (as in prior work), unless stated otherwise.\looseness=-1

Following the conditional setup in \citet{kipf2021conditional}, the initial state of the slots is obtained by encoding bounding boxes corresponding to the objects in the first frame; thus providing the model with rough cues of which objects to bind to initially. For the unconditional experiments, we initialize slots using an equal amount of learnable parameter vectors. In experiments that use optical flow, we convert the 2D flow signal to three RGB channels following prior work~\citep{yang2021self}. As described above, we apply a log-transform to the (sparse) depth signal (incremented by 1 to avoid underflow). We train our model to minimize the squared error (L2 loss) between the predicted and ground-truth targets in pixel space. 
We implement SAVi++ in JAX~\citep{jax2018github} using the Flax~\citep{flax2020github} neural network library. Training SAVi++ on a single MOVi dataset on 8 TPUv4 chips with 32GiB memory each takes approximately two days for 500k training steps.

\section{Model details}
\subsection{SAVi++}

Our architecture building blocks are similar to that of the SAVi model from \citet{kipf2021conditional} with all the parameters shared across time steps. \modelname uses exactly the same parameters for the slot initializer and decoder as the SAVi model (except for \modelname HR where we use an additional $5\times5$ ConvTranspose layer with stride 2 and 64 channels to account for the higher resolution frame size). Below we list the details and hyper parameters of all the modules of \modelname:

\paragraph{Encoder} We used a ResNet-34~\citep{he2016deep} backbone with modified root convolutional layer that has $1\times1$ stride (except for \modelname HR that uses a root stride of $2\times2$). For all layers, we replaced the batch normalization operation by group normalization~\citep{wu2018group}. We used a linear positional encoding identical to that used in Slot Attention~\citep{locatello2020object} with horizontal and vertical coordinates normalized to $[-1, 1]$ range. These coordinates were then projected to the same size of the ResNet feature maps using a learnable linear layer. Finally, the ResNet features and positional encoding are combined by an addition operation. Following the backbone, the frame features are  projected to $64$ embedding dimensions by a linear layer followed by ReLU activation and then fed to a transformer network with 4 transformer blocks, except for \modelname unconditional Waymo Open models where we found that ResNet features alone produced clearer segmentations.
This is likely due to the stronger CNN image prior being beneficial with less supervision~\citep{ulyanov2018deep}. Each transformer block uses a multi-head dot-product attention from \cite{vaswani2017attention} with pre-normalization~\citep{xiong2020layer} and 4 attention heads. For each attention head, the query/key/value embedding size was set to 16. The output of each block is then processed by a residual feed-forward block with pre-normalization, using an MLP with a single hidden layer of 1024 hidden units and ReLU activation function.

\paragraph{Corrector/Predictor} Similar to SAVi~\citep{kipf2021conditional}, we use 1 iteration for the Slot Attention corrector module. We increase the corrector query/key/value size to 256 compared to 128 embedding size used in SAVi. For the predictor, we similarly increased the query/key/value projection size to 256, and the MLP hidden layer size to 1024. We found that the larger embedding sizes for the corrector and predictor increased \modelname mIoU by a few percentage points.

\paragraph{Decoder} Our decoder follows that of SAVi~\citep{kipf2021conditional} with two exceptions: For Waymo Open, we use a larger spatial broadcast grid of $8\times 12$ as video frames have $128\times 192$ resolution. For SAVi++ HR, we use frames of a higher resolution of $256\times 384$ and add an additional $5\times5$ ConvTranspose layer with stride 2 and 64 channels to account for the higher resolution frame size.
The decoder otherwise uses four $5\times5$ ConvTranspose layers with stride 2 and 64 channels, followed by ReLU activations.

\paragraph{Initializer} Similar to SAVi, we consider two initializers \emph{conditional} and \emph{unconditional} to set the initial state of the models' $K$ slots. For the conditional case, the initializer mapped each of $K$ bounding boxes (represented using 4D coordinates) via a trainable MLP to set the $D$ dimensional state of the corresponding slot. For the unconditional case, rather than associating a specific bounding box in a specific video to a slot, we learn $K$ $D$-dimensional initial slot states to be used in all videos.

\paragraph{Data augmentation} As discussed in the main text, an Inception style~\citep{szegedy2015going} random crop is used for data augmentation. Crops are afterwards resized to the target resolution ($128\times128$ for MOVi and $128\times192$ for Waymo Open, unless otherwise mentioned). We ensure that the cropped view covers at least $20\%$ of the original frame in the MOVi datasets and $75\%$ in Waymo Open. We found that more aggressive cropping was helpful for the synthetic MOVi datasets and less aggressive cropping worked best on Waymo Open. We take care to handle depth maps and flow fields during this operation as explained next for each of the datasets. \textit{MOVi:} Say the crop size is $h \times w$, then optical flow fields are cropped, resized, and then rescaled by $\left[ \frac{128}{w}, \frac{128}{h} \right]$. Dense depth maps are simply cropped and resized. \textit{Waymo Open:} Sparse LiDAR point clouds are projected to 2D and retained as tuples (2D point, lidar-range) throughout the data augmentation pipeline. The affine transformation equivalent to the crop and resize operation is computed and applied to these 2D points. As a consequence, several points will fall outside the cropped frame. These points are discarded when projecting them into a depth image after data augmentation is complete.

\subsection{Baselines}

\paragraph{SAVi} For the SAVi baseline, we use the best-performing model variant described in~\citet{kipf2021conditional}, i.e.~SAVi trained with a Resnet34 backbone. Different from SAVi++, this baseline does not use depth prediction (it only predicts optical flow), does not use data augmentation, and does not use a transformer encoder after the convolutional backbone. We choose the same hyperparameters as described in SAVi~\citep{kipf2021conditional}.

\paragraph{SIMONe}
This baseline~\citep{kabra2021simone} is a non-autoregressive model for encoding short video clips of fixed lengths into a set of latent object variables (fixed across time) and a per-frame global latent variable. Note that SIMONe cannot be applied auto-regressively and has to be applied to the same sequence length at both training and test time. SIMONe uses a CNN encoder per frame followed by a transformer encoder that is applied across frames to finally obtain object and frame latent variables by pooling transformer tokens across time and space, respectively. The model is trained by reconstructing input frames using a form of a spatial broadcast decoder and additionally uses a KL-based regularizer on the latent variables. We use a JAX~\citep{jax2018github} reimplementation of the SIMONe model for which we verified that it reproduces results mentioned in the paper on the CATER~\citep{girdhar2019cater} dataset. We train SIMONe on sequences of 6 frames (same as SAVi++) at a resolution of $128\time192$. We subsample reconstruction targets by a factor of 4 (see \citet{kabra2021simone} for details). Other hyperparameters are chosen as follows: reconstruction loss scale $\alpha=0.2$, pixel likelihood scale $\sigma_x=0.08$, object latents KL loss weight $\beta_o=\mathrm{1e-5}$, and frame latents KL loss weight $\beta_f=\mathrm{1e-4}$. We encode frames using a 4-layer CNN with 128 channels, $(4, 4)$ kernel size and $(2, 2)$ stride. Each transformer uses 4 layers, 5 heads, a qkv-size of 64 per head, and an MLP hidden layer size of 1024. Latents are of size 32. The decoder uses an MLP with 5 hidden layers of 512 units. To train SIMONe with sparse depth targets, we replace the RGB target signal with the LiDAR-based depth signal and only compute the loss for pixels that have a depth signal.

\paragraph{CRW} Contrastive Random Walks (CRW)~\citep{jabri2020space} is a cycle consistency based self-supervised learning method for learning grid structured latent representations. After pre-training, a simple label propagation scheme can be applied on these latent representations to obtain tracking behavior. This typically requires segmentation labels for the objects of interest in the first frame. The method tracks these objects and outputs segmentation masks for them over subsequent frames. In order to use this method with only bounding box conditioning in the first frame, we flood fill boxes into rectangular masks and propagate those instead. Overlap between boxes is resolved based on the box order.

We adopted their training and evaluation best practices. We pre-trained stride-8 ResNet backbones~\citep{he2016deep} using their publicly available code and propagated labels using the activations output by the second last ResNet stage. For pre-training we tuned edge-dropout, training temperature and for tracking we tuned evaluation temperature independently on each of the three MOVi datasets. We found that, despite our efforts, a ResNet34 backbone was not able to train using the cycle consistency loss. We obtained much better results using a ResNet18 backbone, which is the model we report results for. Optimal hyper-parameters (dropout, training temperature, evaluation temperature) were as follows: MOVi-C $(0.0, 0.001, 0.5)$, MOVi-D $(0.05, 0.001, 0.5)$, MOVi-E $(0.05, 0.001, 0.5)$. Other relevant hyper-parameters are: training clip length $(6)$, frame-skip $(1)$, batch size $(16)$, learning rate $(0.0001)$, training epochs $(125)$ with a learning rate drop after the $100^{th}$ epoch.

\paragraph{Bounding box copy}
We simply repeat the bounding boxes of the objects visible in the initial frame for the rest of the video sequence. 
To obtain pixel-level segments for computing metrics, such as FG-ARI, we `render' the entire bounding box as a segment in pixel space.
Bounding boxes are rendered in the same order as they were provided in the initial frame, such that later boxes take precedence when multiple of them cover the same pixel location.

\paragraph{Learned bounding box propagation}

In this baseline, we use the SAVi++ initializer and predictor (without encoder, corrector, or decoder) to learn a bounding box propagation model. The model receives (just as in SAVi++) bounding boxes for all objects in the first frame of the video, which are passed to the initializer to learn initial slot representations. Afterwards, the predictor learns a mapping of slots at time step $t$ to slots at time step $t+1$. We train the model by reading out individual slot representations at each time step using an MLP with a single hidden layer of 256 units that predicts the corner coordinates (top-left and bottom-right) of the bounding box associated with a slot at a particular time step, supervised using ground-truth bounding boxes. We use the Huber~\citep{huber1964robust} loss (L2 loss between $[-1, 1]$ and L1 loss outside of this interval) to train the model. If an object is not visible or present, its bounding box is set to $[0, 0, 0, 0]$ in the ground-truth target.

\paragraph{K-Means clustering of flow/depth}
To evaluate how much information about instance segmentation can be exctracted directly from the depth and optical flow modalities, we evaluate a k-Means clustering baseline on videos from MOVi and Waymo Open.
For that purpose we treat each pixel of a video as a datapoint, each with 7 dimensions: one for log-depth ($\log 1 + d$), three for optical flow converted to RGB (only for MOVi), two for linear position encoding, and one for time. 
All dimensions are normalized to the range of $[0, 1]$. 
For Waymo Open we discard all points that do not have an associated depth value. 
To make it as comparable as possible to the conditional setup of SAVi++, we set $k$ to the ground-truth number of objects plus one for the background, and initialize each cluster-center to the average value of points within the first-frame bounding box of each object.
The background cluster is initialized to the average value of all points in the first frame.
K-Means is then run until convergence, and we evaluate the resulting cluster-assignments using mIoU and FG-ARI scores for MOVi, and by computing the normalized distance of the center of mass of each segment to the corresponding bounding box for Waymo Open.

\paragraph{Supervised baseline} To estimate how much headroom there is in terms of tracking performance given our model architecture, we train a variant of the SAVi++ model where we replace the depth decoder with a bounding-box prediction head. Instead of self-supervised training using depth prediction, this model is trained to directly predict object bounding boxes at every time step given the slots of the model. This is similar to a TrackFormer~\citep{meinhardt2021trackformer} model, but we instead use the SAVi++ architecture and we train in a conditional setting (i.e.~initial first-frame bounding boxes are provided as slot initialization), which means we can train the model without using any form of matching. Similar to the learned bounding box propagation baseline, we train the model by reading out individual slot representations at each time step using an MLP with a single hidden layer of 256 units that predicts the corner coordinates (top-left and bottom-right) of the bounding box associated with a slot at a particular time step, supervised using ground-truth bounding boxes. We apply a Huber~\citep{huber1964robust} loss (L2 loss between $[-1, 1]$ and L1 loss outside of this interval) for training. If an object is not visible or present, its bounding box is set to $[0, 0, 0, 0]$ in the ground-truth target.

\section{Datasets}

We used the synthetic Multi-Object Video (MOVi) datasets introduced in Kubric~\citep{kubric2021}. The Kubric dataset generation pipeline is available under an Apache 2.0 license. For real-world experiments, we used the Waymo Open dataset~\citep{sun2020scalability}. The Waymo Open dataset is licensed under the Waymo Dataset License Agreement for Non-Commercial Use (August 2019): \url{https://waymo.com/open/terms}. 

Dataset details are summarized in the following:
\begin{itemize}[leftmargin=*]
    \item \textbf{MOVi-C}: uses approximately 380 high-resolution HDR photos as backgrounds and three to ten dynamic objects obtained from a set of 1028 3D-scanned everyday objects \cite{gso2020}, representing various household objects. The camera in this dataset has a random pose, yet the camera pose is static across the video sequence. Each video is sampled at 12 frames per second (fps). We trained our models on randomly sampled sequences of 6 frames and evaluate on sequences of 24 frames. We trained our models on 9.75k training set videos, and evaluated models on 250 evaluation set videos. Model tuning was performed on a separately generated set of 250 videos.
    \item \textbf{MOVi-D}: has a similar camera setting as MOVi-C, but adds more objects, the majority of which are initialized to be static. This dataset includes one to three dynamic objects and 10 to 20 static objects in each video sequence. We used the same frame rate and train/evaluation sequence lengths as for MOVi-C. We trained our models on 9.75k training set videos, and evaluated models on 250 evaluation set videos. Model tuning was performed on a separately generated set of 250 videos.
    \item \textbf{MOVi-E}: adds additional complexity compared to MOVi-D by introducing random linear camera movement throughout the video sequence. We used the same frame rate and train/evaluation sequence lengths as for MOVi-C. We trained our models on 9.75k training set videos, and evaluated models on 250 evaluation set videos. Model tuning was performed on a separately generated set of 250 videos.
    
    \item \textbf{Waymo Open}: contains high-resolution video sequences with frame size of $1280\times1920$. We solely use videos recorded from the front camera of the car. We down-sampled the videos to $128\times192$ (or $256\times384$ for \modelname HR). The dataset consists of 798 train and 202 validation scenes of 20s video each, sampled at 10 fps. The dataset includes also 2D bounding box annotations, which we used for the conditional experiments and to compute the B.~mIoU evlauation metrics. The Waymo Open dataset further includes LiDAR data collected from five LiDARs; one mid-range LiDARs placed on top of the car and four short-range LiDARs placed front, left, right, and rear. The LiDAR data is used to compute sparse depth targets as discussed in the Methods section. To slightly simplify the task as we train on lower-resolution frames, we discard any bounding box labels in Waymo Open which cover an area of $0.5\%$ or less of the first sampled video frame, both during training and testing.
\end{itemize}

\section{Metrics}

In the following, we give a detailed overview of the metrics used for each of the datasets.

\subsection{MOVi}

On the MOVi datasets~\citep{kubric2021} we have access to ground-truth pixel-level segmentations, which lets us directly measure the quality of the learned segmentations using the same segmentation metrics as in prior work.
Note that SAVi and SAVi++ are trained in a conditional setting where we initialize slots using ground-truth bounding box information in the first frame. 
Because of this, we will only measure metrics from the second frame onward.

\textbf{Foreground Adjusted Rand Index (FG-ARI)}\quad
A permutation-invariant clustering similarity metric frequently used for evaluating scene decomposition quality~\citep{rand1971objective,hubert1985comparing}. It compares discovered segmentation masks with ground-truth masks while ignoring any pixels that belong to the background. It is sensitive to temporal consistency of masks, but insensitive to their ordering.

\textbf{Mean Intersection over Union (mIoU)}\quad
A standard segmentation metric for measuring the quality of predicted segments. Our implementation is identical to the semi-supervised DAVIS challenge Jaccard-Mean metric for video~\citep{caelles2019the,ponttuset2017the}. We note that this implementation is sensitive to the correct ordering of masks, i.e.~it also measures whether models used the conditioning signal (here, first-frame bounding boxes) correctly.

Model selection on MOVi was done mainly using mIoU. On the one hand to avoid learned segments bleeding into the background overly much, and on the other hand to ensure that the bounding box initialization was properly utilized.

\subsection{Waymo Open}

On the Waymo Open dataset we only have ground-truth bounding boxes available, which necessitates an alternative set of metrics for measuring quantitative performance.
Similar to before, because of conditioning, we will only measure metrics from the second frame onward.

\textbf{Center-of-Mass (CoM) distance}\quad 
This tracking metric measures the average Euclidean distance between the centroid of the predicted segmentation masks and the centers of the ground-truth bounding boxes.
The former are obtained by computing the geometric mean of the 2D coordinates associated with the pixels belonging to a segment, where we exclude pixels that do not have a valid LiDAR point associated with them (this is similar to how we compute the loss during training).
To allow for comparable CoM distance across multiple resolutions, we report the distance normalized by the maximum achievable distance in the video frame (length of the diagonal).
Objects that are fully occluded in a frame (or have disappeared) are excluded from the computation.
In the conditional case (i.e.~with bounding box information provided to the model in the first frame) we use the order of the provided bounding boxes to compute the metric between each slot and ground-truth bounding box. 
In the unconditional case, we use Hungarian matching to associate entire bounding box tracks with decoded object masks and we assign a penalty of 1 (maximum CoM distance) for all empty segments.

\textbf{Bounding Box Recall (B. Recall)}\quad
This metric measures the fraction of cases where any sort of segment is predicted when a valid ground-truth box exists.
It serves a complement to CoM distance when no matching is used and empty segments are not considered. In the case of unconditional evaluation using Hungarian matching, we incorporate a matching penalty of the maximum possible CoM distance for empty segments and thus do not separately report segment recall.

\textbf{Bounding Box mIoU (B.~mIoU)}\quad 
This metric is the bounding box analog of the segmentation mIoU discussed above. Given corresponding predicted and ground-truth bounding box tracks, their per-frame intersection-over-union is computed and averaged over time exactly as in the average IoU metric of the TAO benchmark~\cite{dave2020tao}. Predicted bounding box tracks are obtained using a per-slot readout MLP, with one hidden layer of 256 units. This is jointly trained with the \modelname model  by minimizing the Huber loss~\citep{huber1964robust} between predictions and $[0, 1]$ normalized ground-truth box coordinates. A stop-gradient is used to prevent these loss gradients from propagating back into \modelname. Objects that are fully occluded across the entire video sequence are excluded from the computation.

We initially conducted model selection on Waymo Open using a combination of B.~mIoU and a heuristic metric to measure what fraction of the pixels belonging to a predicted segment are inside the associated ground-truth bounding box. During the final stages of development, we primarily focused on the B.~mIoU metric since it is analogous to mIoU on MOVi.